\documentclass[twoside]{article}
\usepackage{titletoc}
\usepackage{appendix}
\usepackage{multirow}
\usepackage{adjustbox}
\usepackage{booktabs}
\usepackage{threeparttable}
\usepackage{graphicx}
\usepackage{rotating}
\usepackage{booktabs} % For better table lines
\usepackage{caption}
\usepackage{algorithm}
\usepackage{algorithmic}
\usepackage{threeparttable}
\usepackage{multicol}
\usepackage{multirow}
\usepackage{booktabs}
\usepackage{threeparttable}
\usepackage{graphicx}
\usepackage{multirow}
\usepackage{rotating}
\usepackage{subfig}
\usepackage[round]{natbib}

\usepackage[colorlinks,bookmarksopen,bookmarksnumbered,citecolor=green, linkcolor=red, urlcolor=green]{hyperref}
\usepackage{hyperref}
\graphicspath{{figures/}}
\usepackage[accepted]{arXiv_double}
\usepackage{amsmath}  
\usepackage{amssymb} 
\begin{document}

% If your paper is accepted and the title of your paper is very long,
% the style will print as headings an error message. Use the following
% command to supply a shorter title of your paper so that it can be
% used as headings.
%
%\runningtitle{I use this title instead because the last one was very long}

% If your paper is accepted and the number of authors is large, the
% style will print as headings an error message. Use the following
% command to supply a shorter version of the authors names so that
% they can be used as headings (for example, use only the surnames)
%
%\runningauthor{Surname 1, Surname 2, Surname 3, ...., Surname n}

\twocolumn[

\aistatstitle{FreqMoE: Enhancing Time Series Forecasting through Frequency Decomposition Mixture of Experts}

\aistatsauthor{ Ziqi Liu }

\aistatsaddress{ Xi'an Jiaotong Liverpool University } ]

\begin{abstract}
Long-term time series forecasting is essential in areas like finance and weather prediction. Besides traditional methods that operate in the time domain, many recent models transform time series data into the frequency domain to better capture complex patterns. However, these methods often use filtering techniques to remove certain frequency signals as noise, which may unintentionally discard important information and reduce prediction accuracy. To address this, we propose the Frequency Decomposition Mixture-of-Experts (FreqMoE) model, which dynamically decomposes time series data into frequency bands, each processed by a specialized expert. A gating mechanism adjusts the importance of each output of expert based on frequency characteristics, and the aggregated results are fed into a prediction module that iteratively refines the forecast using residual connections. Our experiments demonstrate that FreqMoE outperforms state-of-the-art models, achieving the best performance on 51 out of 70 metrics across all tested datasets, while significantly reducing the number of required parameters to under 50k, providing notable efficiency advantages. Code is available at: \href{https://github.com/sunbus100/FreqMoE-main}{https://github.com/sunbus100/FreqMoE-main}
\end{abstract}
\section{INTRODUCTION}

The task of time series forecasting aims to predict the future based on known past time series data and has broad applications in fields such as finance\citep{ariyo2014stock}, weather forecasting\citep{zhu2023weather2kmultivariatespatiotemporalbenchmark}, energy consumption\citep{zufferey2017forecasting}, and healthcare management\citep{zeroual2020deep}. Traditional statistical methods, such as the Autoregressive Integrated Moving Average \citep{nelson1998time} and exponential smoothing\citep{gardner1985exponential}, were once widely used for time series analysis. However, these methods often perform poorly when dealing with nonlinear, non-stationary, and multi-sequence real-world datasets\citep{li2012}.

In recent years, to address the limitations of traditional models in capturing complex patterns and dynamic features in time series, numerous studies have proposed sophisticated models and methodologies based on MLP\citep{das2024longtermforecastingtidetimeseries,wang2024timemixerdecomposablemultiscalemixing,zeng2022transformerseffectivetimeseries} and transformer architectures\citep{zhou2021informerefficienttransformerlong,zhou2022fedformerfrequencyenhanceddecomposed,wu2022autoformerdecompositiontransformersautocorrelation,nie2023timeseriesworth64}, which have achieved significant improvements over traditional methods. Despite their success, these models primarily operate in the time domain and may not fully leverage the frequency characteristics of time series data, such as seasonality and cyclic patterns\citep{box2015time}.

As a result, more and more models are shifting towards frequency domain analysis\citep{liu2023koopalearningnonstationarytime,zhou2022fedformerfrequencyenhanceddecomposed,wu2023timesnettemporal2dvariationmodeling}. By utilizing techniques such as Fast Fourier Transform (FFT), time series data can be decomposed into their respective frequency components, making it easier to analyze periodic behaviors and trends that are difficult to observe in the time domain. Although some models, such FITS\citep{xu2024fitsmodelingtimeseries}, have employed frequency domain analysis for time series forecasting, but it relies on fixed filtering techniques. We believe that applying fixed filters indiscriminately across different datasets is unreasonable. Instead, the weights for different frequency bands should be dynamically assigned based on the frequency characteristics of the data to prevent the loss of important information and improve forecasting accuracy.

In this paper, we propose a novel Frequency Decomposition Mixture-of-Experts model. Our approach leverages the frequency domain representation of time series data by decomposing the input into different frequency bands, with each expert network focusing on a specific frequency range. A gating network dynamically assigns weights to each expert based on the frequency magnitude, enabling a data-driven combination of expert outputs. The aggregated output is then fed into a prediction architecture consisting of multiple deep residual blocks, achieving end-to-end time series forecasting.

Our primary contributions include:
\begin{enumerate}
    \item We introduce a frequency decomposition MoE module where each expert network focuses on a specific frequency range of the input data. This specialization enables the model to capture intricate patterns associated with different frequency components.
    
    \item We develop a gated network that operates on the frequency magnitude spectrum, dynamically weighting the contribution of each expert according to the frequency properties of the input. This mechanism allows the model to more fully utilize all the information and extract the dominant frequency patterns.
    
    \item A prediction structure combining the frequency upsampling technique with the stacked depth residual module is proposed, which together with the frequency decomposition MoE module forms FreqMoE, a powerful forecast model, achieves consistent state-of-the-art performance across multiple domains.
\end{enumerate}

\section{RELATED WORK AND MOTIVATION}

\subsection{Long-horizon Time-series Forecasting Models}

Transformer-based models were originally developed for natural language processing\citep{vaswani2017attention}, but have been used for long-term time series forecasting due to their ability to capture long-term dependencies through a self-attention mechanism. For example, the earliest Transformer-based model, Informer\citep{zhou2021informerefficienttransformerlong}, solves the problem of squared time complexity growth by using the self-attention extraction mechanism, as well as Autoformer\citep{wu2022autoformerdecompositiontransformersautocorrelation}, which uses autocorrelation decomposition, and FEDformer\citep{zhou2022fedformerfrequencyenhanceddecomposed}, which employs frequency augmentation and sequence decomposition, and PatchTST\citep{nie2023timeseriesworth64}, which is the most recent model that introduces the idea of patch. These models have been shown to have good results in complex time series prediction. 

Despite the success of transformer-based models in long-term time series forecasting, \citep{zeng2022transformerseffectivetimeseries}, still gave doubts about the effectiveness of transformer-based models and proposed a Dlinear model consisting of only one simple linear layer, which proved the simplicity and efficiency of MLP-based models. Similarly, the subsequently proposed TimeMixer model\citep{wang2024timemixerdecomposablemultiscalemixing} also uses only multiple MLP layers combined with a downsampling mechanism to capture temporal dependencies, showing that a well-designed MLP model can match or exceed transformer in prediction tasks. 

\subsection{Motivation}

With the continuous development of the field, some models are gradually shifting from time-domain prediction to frequency-domain prediction, such as FEDformer use frequency domain enhancement module combined with transformer for prediction\citep{zhou2022fedformerfrequencyenhanceddecomposed}, TimesNet decompose time series into multiple periods by Fourier tansforms \citep{wu2023timesnettemporal2dvariationmodeling}, and FITS predict with complex linear layer after apply a low-pass filter\citep{xu2024fitsmodelingtimeseries}. However, many frequency-domain prediction models now treat high-frequency components as noise that should be discarded\citep{xu2024fitsmodelingtimeseries,zhou2022film}, but directly recognizing high-frequency components as noise in the absence of a prior knowledge may result in the loss of important information, thus reducing the prediction accuracy\citep{zhang2024frequenciescreatedequaltowardsdynamic}.

To address this problem, we believe that the fixed frequency components in the original sequence should not be directly deleted. Instead, the weights of each frequency should be dynamically adjusted according to the data, and different frequencies should be weighted and combined\citep{zhang2024frequenciescreatedequaltowardsdynamic}. This can make full use of all the information and accurately extract the key frequency patterns, thus realizing more efficient prediction. Inspired by the stackable residual architecture\citep{oreshkin2020nbeatsneuralbasisexpansion}, we argue that we can exploit the properties of frequency domain signals by upsampling the sequence while reconstructing the backtracking window and making predictions. This allows the model to be iteratively optimized based on the residuals of the backtracking sequence, progressively capturing complex patterns at different frequencies.

\section{PRELIMINARIES}
\subsection{Problem Definition}
Time series forecasting aims to predict future values of a sequence based on historical observations. Formally, let \( X \in \mathbb{R}^{c \times s} \) represent the input sequence, where \( c \) is the number of channels or features, and \( s \) denotes the number of historical time steps. The goal is to learn a mapping function \( f \) such that \( f(X) = Y \), where \( Y \in \mathbb{R}^{c \times p} \) is the predicted sequence of future values over \( p \) time steps. This mapping should minimize a predefined loss function \( \text{loss}(Y, \hat{Y}) \), with \( \hat{Y} \) being the ground truth future values.

Generally, the forecasting performance of the model is typically evaluated using the mean squared error (MSE) and the mean absolute error (MAE) as loss functions. These functions are defined as:
\begin{equation}
\text{MSE} = \frac{1}{n} \sum_{i=1}^{n} (Y_i - \hat{Y}_i)^2    
\end{equation}
\begin{equation}
\text{MAE} = \frac{1}{n} \sum_{i=1}^{n} |Y_i - \hat{Y}_i|    
\end{equation}
\subsection{Fast Fourier Transform(FFT)}
The Discrete Fourier Transform (DFT) converts discrete-time signals from the time domain to the complex frequency domain, while the Fast Fourier Transform \citep{brigham1967fast} significantly accelerates the computation of the DFT through an efficient algorithm. For real-valued input signals, the Real Fast Fourier Transform (rFFT) is typically employed, transforming $N$ real input values into $N/2 + 1$ complex numbers, thereby effectively representing the signal in the complex frequency domain.

\section{METHODOLOGY}
\subsection{Overview}

\begin{figure*}
\centering 
\includegraphics[width=1.0\textwidth]{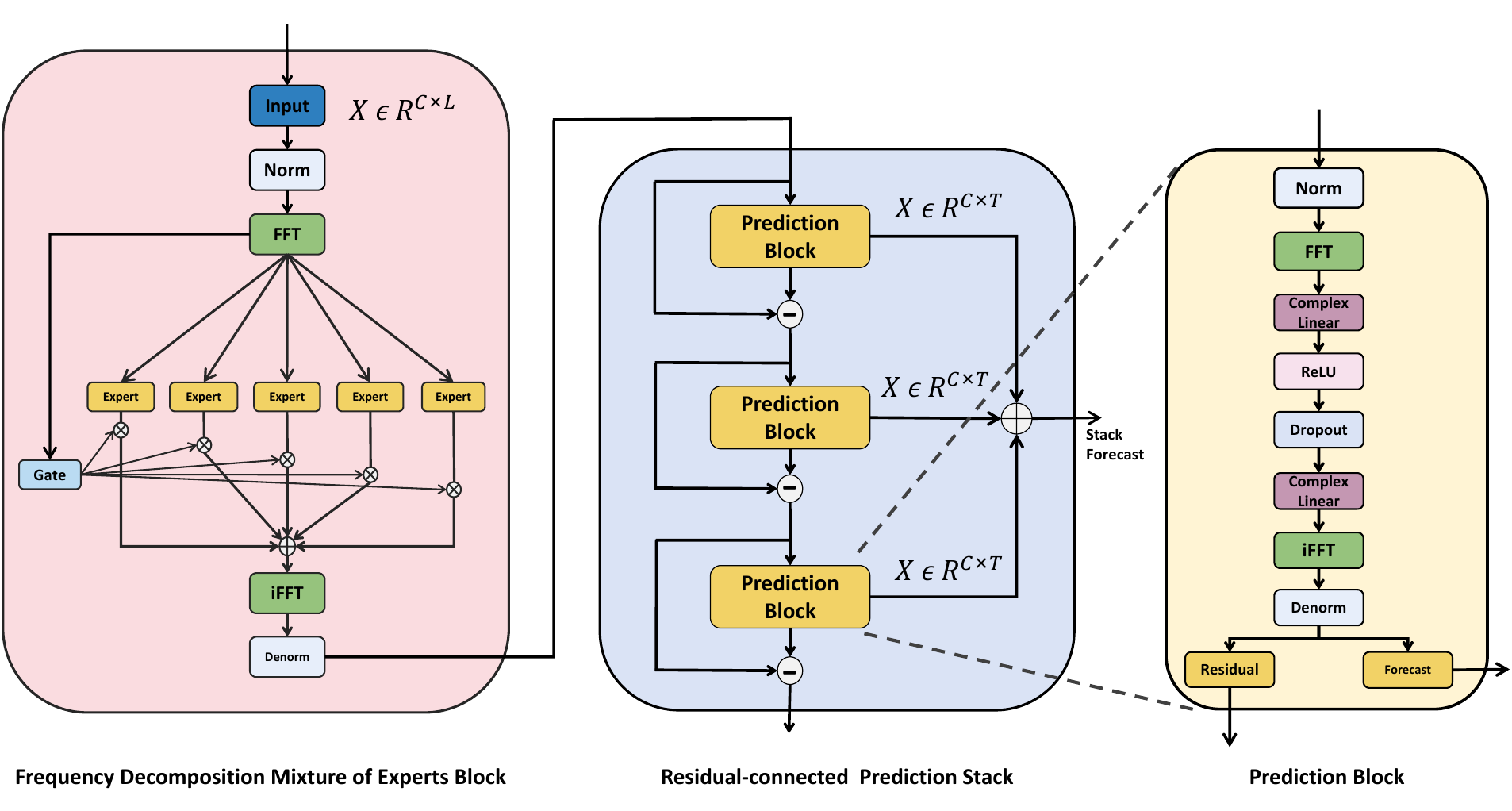} 
\caption{The complete structure of the FreqMoE model consists of a Frequency Decomposition Mixture of Experts Block, which decomposes and processes the input time series before reconstruction, and a Frequency Domain Prediction Stack, which is composed of multiple prediction blocks connected through residual connections. This model dynamically captures patterns in the sequence by leveraging the relationships and frequency characteristics across different frequency bands of the input data, enabling accurate predictions.} 
\label{fig:structure} 
\end{figure*}

Motivated by exploration of frequency-domain time series forecasting and raw data filtering challenges in section 2, we introduce a deep architecture based on frequency decomposition. As illustrated in Figure \ref{fig:structure}, the architecture comprises a Frequency Decomposition Mixture-of-Experts module and a Frequency Domain Prediction module, the latter constructed by stacking multiple prediction blocks with residual connections. Detailed descriptions of the Frequency Decomposition Expert Network and Prediction modules are provided in Sections 4.2 and 4.3, respectively.

The input time series $\mathbf{x} \in \mathbb{R}^{B \times C \times L}$, where $B$ is the batch size, $C$ the number of channels, and $L$ the sequence length, is first normalized using mean subtraction and variance scaling, resulting in zero mean and unit variance data. Then, the normalized data is transformed into the frequency domain using the Fast Fourier Transform (FFT), resulting in
\begin{equation}
\mathbf{X}_{f} = \text{FFT}(\mathbf{x}) \in \mathbb{C}^{B \times C \times L_{f}},
\end{equation}
where $L_f = \frac{L}{2} + 1$ is the length of the frequency domain representation. The frequency components are then divided into $N_{\text{expert}}$ disjoint regions, with each expert handling a specific frequency band. The gating network, $G(\mathbf{X}_f)$, takes the magnitude of the frequency components, $\|\mathbf{X}_f\|$, as input and generates gating scores $\mathbf{g} \in \mathbb{R}^{B \times N_{\text{expert}}}$ using a softmax activation:
\begin{equation}
\mathbf{g} = \text{Softmax}(G(\|\mathbf{X}_f\|)).
\end{equation}
The expert outputs are combined using frequency-domain gating scores, then transformed back to the time domain with inverse FFT and denormalized. This module enables the model to capture information from each frequency band and manage short- and long-term temporal patterns effectively.

The output from the Frequency Decomposition Mixture of Experts module is processed through several frequency domain prediction blocks, each containing two simple complex-valued linear layers. Each block upsamples the Fourier-transformed representation $\mathbf{X}_f$, reconstructs historical sequences, and predicts future ones. After an inverse Fourier transform and denormalization, the residual between the output and input sequences is used as input for the next block, progressively refining the predictions:
\begin{equation}
\mathbf{x}^{(i+1)} = \mathbf{x}^{(i)} - \hat{\mathbf{x}}^{(i)}
\end{equation}
The final predicted sequence is obtained by accumulating the predictions of each block.

\subsection{Frequency Decomposition Mixture of Experts Block}

The Frequency Decomposition Mixture of Experts (MoE) module is one of the core components of our model. The input time series $\mathbf{X} \in \mathbb{R}^{B \times C \times L}$, where $B$ is the batch size, $C$ is the number of channels, and $L$ is the sequence length, is first transformed into the frequency domain using the Fast Fourier Transform (FFT), resulting in a frequency domain sequence $\mathbf{F}(\mathbf{X}) \in \mathbb{C}^{B \times C \times F}$ of length $F = \frac{L}{2} + 1$. In the frequency domain, the sequence is represented in complex form, where each component corresponds to the amplitude and phase of a specific frequency.

To fully utilize the information across all frequency bands, we introduce expert networks, with each expert responsible for processing a specific frequency range. we propose to learn the frequency band boundaries in an end-to-end manner. This allows the model to adaptively focus on the most informative frequency bands.

We parameterize the frequency band boundaries using learnable parameters $\boldsymbol{\theta} = \{\theta_1, \theta_2, \dots, \theta_{N-1}\}$, where $N$ is the number of experts. Each $\theta_i$ is a scalar value. We apply the sigmoid function to map these parameters to the interval $(0, 1)$:
\begin{equation}
    \tilde{b}_i = \sigma(\theta_i) = \frac{1}{1 + e^{-\theta_i}}, \quad i = 1, 2, \dots, N-1.
\end{equation}
To ensure that the frequency bands are non-overlapping and collectively cover the entire frequency range, we sort the normalized boundaries in ascending order, including \(0\) and \(1\) as the initial and final boundaries:
\begin{equation}
    \{b_0, b_1, \dots, b_N\} = \text{Sort}\left(\{0, \tilde{b}_1, \dots, \tilde{b}_{N-1}, 1\}\right),
\end{equation}
where \(b_0 = 0\) and \(b_N = 1\). These normalized boundaries are then mapped to actual frequency indices based on the frequency domain length \(F\). For the \(i\)-th expert \(E_i\), the responsible frequency range is defined as:
\begin{equation}
    \text{Range}_i =
    \begin{cases}
        [f_{i-1}, f_i), & \text{for } i = 1, \dots, N-1, \\
        [f_{N-1}, F], & \text{for } i = N.
    \end{cases}
\end{equation}
To ensure that each expert only processes its assigned frequency range, we construct a mask $M_i \in \{0, 1\}^{B \times C \times F}$ for each expert $E_i$, defined as:
\begin{equation}
    M_i(f) = 
    \begin{cases} 
    1, & \text{if } f \in [f_{i-1}, f_i), \quad i = 1, 2, \dots, N-1, \\
    1, & \text{if } f \in [f_{N-1}, F], \quad i = N, \\
    0, & \text{otherwise}.
    \end{cases}
\end{equation}
The frequency components processed by expert $E_i$ are obtained by applying the mask:
\begin{equation}
    \mathbf{F}_i(\mathbf{X}) = M_i \odot \mathbf{F}(\mathbf{X}),
\end{equation}
where $\odot$ denotes element-wise multiplication, and $\mathbf{F}_i(\mathbf{X})$ represents the frequency sub-band processed by expert $E_i$ after applying the mask.

After obtaining output of each expert, the gating network in the Mixture of Experts module dynamically aggregates them based on the frequency domain characteristics of input sequence. Specifically, it computes the magnitude of the frequency representation and averages it across the channel dimension to serve as input for the gating network:
\begin{equation}
    \mathbf{G}(\mathbf{X}) = \frac{1}{C} \sum_{c=1}^{C} \left| \mathbf{F}(\mathbf{X})_c \right| \in \mathbb{R}^{B \times F}.
\end{equation}
This input $\mathbf{G}(\mathbf{X})$ is subsequently passed through a gating network, a simple linear layer, to produce the weight scores for each expert:
\begin{equation}
    \mathbf{W}(\mathbf{X}) = \text{softmax}\left( \text{Linear}\left( \mathbf{G}(\mathbf{X}) \right) \right) \in \mathbb{R}^{B \times N}.
\end{equation}
These weights represent the relative contribution of each expert to the final combined output.

Finally, we compute a weighted sum of all expert outputs in the frequency domain, resulting in the final frequency domain output:
\begin{equation}
    \mathbf{F}_{\text{out}}(\mathbf{X}) = \sum_{i=1}^{N} \mathbf{W}_i(\mathbf{X}) \cdot \mathbf{F}_i(\mathbf{X}) \in \mathbb{C}^{B \times C \times F},
\end{equation}
where $\mathbf{W}_i(\mathbf{X})$ is the weight of expert $E_i$ for each sample.

Once the final frequency domain output is obtained, the inverse Fast Fourier Transform is applied to convert the frequency domain signal back into the time domain:
\begin{equation}
    \mathbf{X}_{\text{out}} = \text{IFFT}\left( \mathbf{F}_{\text{out}}(\mathbf{X}) \right) \in \mathbb{R}^{B \times C \times L}.
\end{equation}
The output $\mathbf{X}_{\text{out}}$ is then passed to subsequent residual connection stacking prediction modules.

\subsection{Residual-connected Frequency Domain Prediction Block}
\subsubsection{Domain Prediction Block}
The purpose of the stackable residual blocks in our model is to iteratively refine the predictions by capturing the residual information that was not explained by previous components. To enable residual connections, each block incorporates two learnable complex-valued linear layers for upsampling, which extends the sequence length to cover both the look-back steps and future time steps. This process is carried out in the frequency domain.

Given an input residual sequence \( r^{(i-1)} \in \mathbb{R}^{c \times s} \), where \( c \) denotes the number of channels and \( s \) represents the sequence length, we first transform \( r^{(i-1)} \) into the frequency domain using the Real Fast Fourier Transform (rFFT):
\begin{equation} 
\mathbf{R}_{\text{freq}}^{(i-1)} = \text{rFFT}(\mathbf{r}^{(i-1)}, \text{dim}=1) \in \mathbb{C}^{c \times (s/2+1)} 
\end{equation}

Here, \( R_{\text{freq}}^{(i-1)} \) consists of complex numbers representing the frequency components of the input signal(\( R_{\text{freq}}^{(i-1)} \)[n] = a + bi). In the frequency domain, complex numbers represent both the amplitude and phase of each frequency component. After Fourier transformation, the signal is transformed into complex numbers, where the real and imaginary parts correspond to the cosine and sine components, respectively. The magnitude (amplitude) of a frequency component is given by \( |z| = \sqrt{a^2 + b^2} \), and the phase (angle) is \( \theta = \arctan \left( \frac{b}{a} \right) \). This representation can fully capture the characteristics of each frequency component.

To predict future time steps, we upsample the frequency components to match the desired output length \( s_{\text{out}} = s + p \), where \( p \) is the number of future time steps to be predicted. Upsampling is done via a two-step complex-valued linear transformation. First, the frequency components pass through a complex linear layer, followed by complex ReLU activation and dropout. The activated components are then processed by a second complex linear layer to generate the final upsampled frequency representation.

The process can be mathematically described as:

\begin{align}
\mathbf{H}^{(i)} &= \mathbf{W}_{\text{up1}}^{(i)} \mathbf{R}_{\text{freq}}^{(i-1)} + \mathbf{b}_{\text{up1}}^{(i)} \in \mathbb{C}^{c \times \left(\frac{s_{\text{out}}}{2}+1\right)}\label{eq:first_linear} \\
\mathbf{H}^{(i)} &= \text{ComplexDropout}\left(\text{ComplexReLU}\left(\mathbf{H}^{(i)}\right)\right)\label{eq:combined} \\
\tilde{\mathbf{R}}_{\text{freq}}^{(i)} &= \mathbf{W}_{\text{up2}}^{(i)} \mathbf{H}^{(i)} + \mathbf{b}_{\text{up2}}^{(i)} \in \mathbb{C}^{c \times \left(\frac{s_{\text{out}}}{2}+1\right)} \label{eq:second_linear}
\end{align}

Where \( \mathbf{W}_{\text{up1}}^{(i)} \in \mathbb{C}^{\left(\frac{s_{\text{out}}}{2} + 1\right) \times \left(\frac{s}{2} + 1\right)} \), and \( \mathbf{W}_{\text{up2}}^{(i)} \in \mathbb{C}^{\left(\frac{s_{\text{out}}}{2} + 1\right) \times \left(\frac{s_{\text{out}}}{2} + 1\right)} \).

By using the complex-valued linear layer, the model can learn to predict the scaling of the magnitude and the shift of the phase for each frequency component, which is crucial for accurate time series forecasting.

After upsampling and transforming the frequency components, we convert them back to the time domain, this results in a time-domain signal that includes the reconstructed look-back sequence as well as predictions for future time steps.

Finally, due to the change in sequence length from \( s \) to \( s_{\text{out}} \) , we adjust the amplitude of the reconstructed signal to ensure proper scaling. This adjustment is achieved by multiplying the signal by the length ratio:
\begin{equation} \hat{\mathbf{y}}^{(i)} = \tilde{\mathbf{R}}_{\text{freq}}^{(i-1)} \times \left( \frac{s_{\text{out}}}{s} \right). \end{equation}
Where \(\hat{\mathbf{y}}^{(i)}\) is the final output of one prediction block.

\subsubsection{Residual Linked Mechanism}
Our model employs a residual learning mechanism where each block iteratively refines the prediction by modeling the residual errors from previous blocks. Let \( r^{(0)} \) denote the initial residual input to the stackable blocks, which is the output from the FreqDecompMoE module. The residual at the \( i \)-th block is defined as:
\begin{equation} \mathbf{r}^{(i)} = \mathbf{r}^{(i-1)} - \hat{\mathbf{y}}^{(i-1)}_{\text{input}}, \end{equation}
Each block takes the current residual \( r^{(i)} \) and produces a prediction \( \hat{y}^{(i)} \) for both the input and future time steps \(\hat{\mathbf{y}}^{(i)} = \text{Block}^{(i)}(\mathbf{r}^{(i)}).\) By stacking multiple residual blocks, the residual error is progressively reduced. The total prediction \( \hat{Y} \) is obtained by accumulating the future predictions from all blocks:
\begin{equation} 
\hat{\mathbf{Y}} = \sum_{i=1}^{N} \hat{\mathbf{y}}^{(i)}
\end{equation}

\section{EXPERIMENT}
\subsection{Main Result}
\begin{table*}[!ht]
  \vspace{-5pt}
  \renewcommand{\arraystretch}{0.9} %行间距
  \centering
  \resizebox{\textwidth}{!}{
  \scalebox{0.8}{
    \begin{threeparttable}
      \small
      \renewcommand{\multirowsetup}{\centering}
      \setlength{\tabcolsep}{1.1pt}
      \caption{Long-term multivariate forecasting results with prediction lengths \( S \in \{96, 192, 336, 720\} \) and fixed lookback length \( T = 96 \). The best averaged forecasting results are in bold and the second-best are underlined. Lower MSE/MAE indicates more accurate predictions.The results of other models are sourced from iTransformer\citep{liu2024itransformerinvertedtransformerseffective}. See Appendix \ref{appendix:full comparison experiment} for the full results.}
      \begin{tabular}{cc|cc|cc|cc|cc|cc|cc|cc|cc|cc|cc}
    \toprule
    \multicolumn{2}{c}{\multirow{1}{*}{Models}} & 
    \multicolumn{2}{c}{\rotatebox{0}{{{FreqMoE}}}} &
    \multicolumn{2}{c}{\rotatebox{0}{{iTransformer}}} &
    \multicolumn{2}{c}{\rotatebox{0}{{PatchTST}}} &
    \multicolumn{2}{c}{\rotatebox{0}{{Crossformer}}}  &
    \multicolumn{2}{c}{\rotatebox{0}{{TiDE}}} &
    \multicolumn{2}{c}{\rotatebox{0}{{{TimesNet}}}} &
    \multicolumn{2}{c}{\rotatebox{0}{{DLinear}}}&
    \multicolumn{2}{c}{\rotatebox{0}{{SCINet}}} &
    \multicolumn{2}{c}{\rotatebox{0}{{FEDformer}}} &
    \multicolumn{2}{c}{\rotatebox{0}{{Autoformer}}} \\
    \cmidrule(lr){3-4} \cmidrule(lr){5-6}\cmidrule(lr){7-8} \cmidrule(lr){9-10}\cmidrule(lr){11-12}\cmidrule(lr){13-14} \cmidrule(lr){15-16} \cmidrule(lr){17-18} \cmidrule(lr){19-20} \cmidrule(lr){21-22}
    \multicolumn{2}{c}{Metric}  & {MSE} & {MAE}  & {MSE} & {MAE}  & {MSE} & {MAE}  & {MSE} & {MAE}  & {MSE} & {MAE}  & {MSE} & {MAE} & {MSE} & {MAE} & {MSE} & {MAE} & {MSE} & {MAE} & {MSE} & {MAE}\\
    \toprule
{ETTm1} && \textbf{0.375} & \textbf{0.396} & 0.407 & 0.410 & \underline{0.387} & \underline{0.400} & 0.513 & 0.496 & 0.419 & 0.419 & 0.400 & 0.406  & 0.403 & 0.407 & 0.485 & 0.481  & 0.448 & 0.452 & 0.588 & 0.517 \\ 
{ETTm2}& &\textbf{0.270} & 0.337 & 0.288 & \underline{0.332} & \underline{0.281} & \textbf{0.326} & 0.757 & 0.610 & 0.358 & 0.404 & 0.291 & 0.333 & 0.350 & 0.401 & 0.571 & 0.537 & 0.305 & 0.349 & 0.327 & 0.371 \\ 
{ETTh1} & & \textbf{0.440} & \textbf{0.429} & 0.454 & \underline{0.447} & 0.469 & 0.454 & 0.529 & 0.522 & 0.541 & 0.507 & 0.458 & 0.450 & 0.456 & 0.452 & 0.747 & 0.647 & \underline{0.440} & 0.460 & 0.496 & 0.487 \\ 
{ETTh2} & & \textbf{0.367} & \textbf{0.397} & \underline{0.383} & \underline{0.407} & 0.387 & 0.407 & 0.942 & 0.684 & 0.611 & 0.550  & 0.414 & 0.427 & 0.559 & 0.515 & 0.954 & 0.723 & 0.437 & 0.449 & 0.450 & 0.459 \\ 
{Exchange} & & \textbf{0.343} & \textbf{0.394} & 0.360 & \underline{0.403} & 0.367 & 0.404 & 0.940 & 0.707 & 0.370 & 0.413 & 0.416 & 0.443 & \underline{0.354} & 0.414 & 0.750 & 0.626 & 0.519 & 0.429 & 0.613 & 0.539 \\ 
{Weather} & &\textbf{0.247} & \textbf{0.276} & \underline{0.258} & \underline{0.279} & 0.259 & 0.281 & 0.259 & 0.315 & 0.271 & 0.320 & 0.259 & 0.287 & 0.265 & 0.317 & 0.292 & 0.363 & 0.309 & 0.360 & 0.338 & 0.382 \\ 
{ECL} & & \underline{0.179} & \textbf{0.270} & \textbf{0.178} & \underline{0.270} & 0.205 & 0.290 & 0.244 & 0.334 & 0.251 & 0.344 & 0.192 & 0.295 & 0.212 & 0.300 & 0.268 & 0.365 & 0.214 & 0.327 & 0.227 & 0.338 \\ 
\bottomrule
  \end{tabular}
    \label{tab:comparison}
    \end{threeparttable}
  }
  }
\end{table*}

We conduct extensive experiments to evaluate FreqMoE on multiple time series forecasting benchmarks(ETTh1, ETTh2, ETTm1, ETTm2, Weather \citep{angryk2020multivariate}, ECL \citep{khan2020towards}, Exchange). Specific descriptions of these benchmarks can be found in Appendix \ref{appendix:datset description}. To ensure a fair comparison, we adopt the experimental settings used in iTransformer\citep{liu2024itransformerinvertedtransformerseffective}. Specifically, the prediction lengths for both training and evaluation are selected from the set $S \in \{96, 192, 336, 720\}$, while maintaining a fixed lookback window of $T = 96$. Apart from these datasets for long-term time series forecasting, we also use the PEMS dataset to test the short-term forecasting performance. Full results and analysis are provided in Appendix \ref{appendix: short term forecast}.

To ensure a comprehensive and fair comparison, we compare FreqMoE against nine state-of-the-art models, covering three mainstream model architectures: Transformer-based PatchTST\citep{nie2023timeseriesworth64}, iTransformer\citep{liu2024itransformerinvertedtransformerseffective},Crossformer\citep{zhang2023crossformer},FEDformer\citep{zhou2022fedformerfrequencyenhanceddecomposed}, Autoformer\citep{wu2022autoformerdecompositiontransformersautocorrelation}, MLP-based DLinear\citep{zeng2022transformerseffectivetimeseries}, TiDE\citep{das2024longtermforecastingtidetimeseries} and TCN-based SCINet\citep{liu2022scinettimeseriesmodeling}, TimesNet\citep{wu2023timesnettemporal2dvariationmodeling}.

Table \ref{tab:comparison} summarizes the performance of FreqMoE on the long-term multivariate prediction task, with the best results highlighted in bold and the next best results underlined. lower MSE/MAE values indicate more accurate predictions. It is noteworthy that our model exhibits excellent performance, achieving the best results in 51 out of 70 benchmark tests, underscoring its robustness and effectiveness. Notably, FreqMoE has shown great dominance especially in long range and low-channel data prediction.

\subsection{Ablation Experiment and Analysis}
\begin{table*}[!ht]
\centering
\renewcommand{\arraystretch}{0.8} 
\caption{Ablation of Frequency Decomposition Module. The best Forecasting results in bold.}
\label{tab:ablation of freqdecomp}
\begin{adjustbox}{max width=\linewidth}
\begin{tabular}{@{}c|cccccccc|cccccccc@{}}
\toprule
Dataset & \multicolumn{8}{c|}{ETTh1} & \multicolumn{8}{c}{Weather} \\ \midrule
Horizon & \multicolumn{2}{c|}{96} & \multicolumn{2}{c|}{192} & \multicolumn{2}{c|}{336} & \multicolumn{2}{c|}{720} & \multicolumn{2}{c|}{96} & \multicolumn{2}{c|}{192} & \multicolumn{2}{c|}{336} & \multicolumn{2}{c}{720} \\ \midrule
Metric & MSE & \multicolumn{1}{c|}{MAE} & MSE & \multicolumn{1}{c|}{MAE} & MSE & \multicolumn{1}{c|}{MAE} & MSE & MAE & MSE & \multicolumn{1}{c|}{MAE} & MSE & \multicolumn{1}{c|}{MAE} & MSE & \multicolumn{1}{c|}{MAE} & MSE & MAE \\ \midrule
w/o Expert & 0.372 & \multicolumn{1}{c|}{0.390} & 0.429 & \multicolumn{1}{c|}{0.424} & 0.478 & \multicolumn{1}{c|}{0.449} & 0.497 & 0.468 & 0.175 & \multicolumn{1}{c|}{0.221} & 0.219 & \multicolumn{1}{c|}{0.261} & 0.275 & \multicolumn{1}{c|}{0.295} & 0.351 & 0.355 \\
Three Expert & \textbf{0.371} & \multicolumn{1}{c|}{\textbf{0.388}} & \textbf{0.426} & \multicolumn{1}{c|}{\textbf{0.422}} & \textbf{0.475} & \multicolumn{1}{c|}{\textbf{0.447}} & \textbf{0.488} & \textbf{0.459} & \textbf{0.168} & \multicolumn{1}{c|}{\textbf{0.215}} &\textbf{0.212} & \multicolumn{1}{c|}{0.254} & 0.270 & \multicolumn{1}{c|}{\textbf{0.291}} & \textbf{0.342} & \textbf{0.345} \\
Five Expert& 0.376 & \multicolumn{1}{c|}{0.395} & 0.435 & \multicolumn{1}{c|}{0.428} & 0.486 & \multicolumn{1}{c|}{0.455} & 0.502 & 0.474 & 0.171 & \multicolumn{1}{c|}{0.217} & 0.214 & \multicolumn{1}{c|}{\textbf{0.253}} & \textbf{0.268} & \multicolumn{1}{c|}{0.292} & 0.343 & 0.347 \\
Eight Expert& {0.372} & \multicolumn{1}{c|}{{0.392}} & {0.426} & \multicolumn{1}{c|}{{0.425}} & {0.480} & \multicolumn{1}{c|}{{0.452}} & {0.492} & {0.471} & {0.172} & \multicolumn{1}{c|}{{0.218}} & {0.215} & \multicolumn{1}{c|}{{0.258}} & {0.273} & \multicolumn{1}{c|}
{{0.292}} & {0.348} & {0.351} \\ \bottomrule
\end{tabular}
\end{adjustbox}
\end{table*}

\subsubsection{Ablation Experiment of Frequency Decomposition MoE Module}
Using experts network to decompose and extract frequency domain signals may be one of the most intriguing components of our model. Therefore, in this section, we will conduct an ablation study to analyze the contributions of the Frequency Decomposition Mixture-of-Experts module. The goal of the ablation is to evaluate the effectiveness of the Frequency Decomposition MoE module in extracting key frequency domain information to enhance the predictive capability of model.

In order to evaluate the performance enhancement effect of the frequency decomposition module, we conducted experiments with the following configurations: 1) no frequency decomposition module; 2) three experts dividing the sequence into three bands; 3) five experts dividing the sequence into five bands; and 4) eight experts dividing the sequence into eight bands. Each configuration was tested on the ETTh1 and Weather datasets according to the settings in Section 5.1. As shown in Table \ref{tab:ablation of freqdecomp}, the three experts models, which basically achieved the lowest MSE and MAE values, proved the effectiveness of the modular feature extraction. However, the performance degraded when five and eight experts were used, we believe that this is mainly due to two reasons: 1) frequency range fragmentation, 2) uneven band coefficients. The specific analysis is in the Appendix \ref{appendix: moe ablation experiment}.

\subsubsection{Effectiveness of Gating Mechanism}

The gating mechanism is the key to the strong generalization capabilities of the decomposition module and to the extraction of temporal patterns,so to validate the effectiveness of it, we compared the performance of model with gating mechanism against those employing only fixed learnable parameters under identical datasets and training conditions (Show in Appendix \ref{appendix:ablation experiment of gating mechanism}). The results demonstrate that models incorporating gating mechanisms significantly outperform those with fixed parameters across all metrics.Also, we observe that the fixed parameter model has essentially the same or even lower loss on the training set as the model with gating mechanism. From this, we believe the improvement is attributable to the ability of gating network to dynamically allocate weights to each frequency band based on input features, allowing the model to adaptively adjust the importance of different frequency bands during the testing phase and accommodate varying frequency distributions. In contrast, models with fixed parameters are unable to adjust frequency band weights according to data characteristics during testing, leading to inaccurate predictions, especially when there is a discrepancy in frequency distributions between the training and testing datasets. Therefore, these findings substantiate the effectiveness of the gating mechanism.

\subsubsection{Effectiveness of Frequency Decomposition MoE Module Improves Other Method as Plugin}
We further investigate the ability of enhancing the performance of other methods since Frequency Decomposition MoE serve as a detachable module. We selected DLinear and PatchTST as baseline models, and the results demonstrate that the performance of all baseline models was improved, as shown in Table \ref{tab:plugin}.

Notably, the module provides greater benefits to simpler models like DLinear. For example, on the ETTh2 dataset with a 720-step horizon, DLinear’s MSE decreases by 16.8\%, while PatchTST only exhibits a 2.1\% reduction. This suggests that Frequency Decomposition MoE compensates for simpler models' weaker capacity to capture temporal dependencies.

Additionally, we observe that the performance gain is more significant for longer forecasting horizons. Specifically, the enhancements at 336 and 720 steps surpass those observed at shorter horizons. This demonstrates the module’s ability to enhance long-range pattern modeling by emphasizing key frequency bands while suppressing noise through dynamically adjusting the weight of different frequency components. To further validate this capability, we designed a synthetic dataset and conducted experiments, with detailed results presented in \ref{appendix:synthetic dataset experiment}.

\renewcommand{\arraystretch}{0.72}
\begin{table*}[!ht]
\setlength{\tabcolsep}{16pt}
\tiny
\centering
\begin{threeparttable}
\caption{Comparison of Frequency Decomposition MoE module improves for different baselines. The best results are in bold.}
\begin{tabular}{c|c|cc|cc|cc|cc}
\toprule

\multicolumn{2}{c}{\scalebox{1.1}{Models}} & \multicolumn{2}{c}{\textbf{DLinear}} & \multicolumn{2}{c}{\textbf{$+$FreqDecompMoE}} & \multicolumn{2}{c}{\textbf{PatchTST}} & \multicolumn{2}{c}{\textbf{$+$FreqDecompMoE}} \\

 \cmidrule(lr){3-4} \cmidrule(lr){5-6} \cmidrule(lr){7-8} \cmidrule(lr){9-10} 

\multicolumn{2}{c}{Metric} & MSE & MAE & MSE & MAE & MSE & MAE & MSE & MAE \\
 
\toprule
\multirow{4}{*}{\rotatebox{90}{ETTh1}}
& 96  & 0.386 & 0.400 &\textbf{0.385} & \textbf{0.398} & 0.414 & 0.419 & \textbf{0.398} & \textbf{0.418} \\
& 192 & 0.437 & 0.432 & \textbf{0.434} & \textbf{0.429} & {0.460} & {0.445} & \textbf{0.426} & \textbf{0.439} \\
& 336 & 0.481 & 0.459 & \textbf{0.478} & \textbf{0.454} & 0.501 & \textbf{0.466} & \textbf{0.494} & {0.471} \\
& 720 & 0.519 & 0.516 & \textbf{0.502} & \textbf{0.500} & \textbf{0.500} & \textbf{0.488} & {0.509} & {0.496} \\
\midrule
\multirow{4}{*}{\rotatebox{90}{ETTh2}}
& 96  & 0.333 & {0.387} & \textbf{0.327} & \textbf{0.382} & \textbf{0.302 }& 0.348 & {0.316} & \textbf{0.347} \\
& 192 &{0.477} &{0.476} & \textbf{0.465} & \textbf{0.468} & 0.388 & \textbf{0.400} & \textbf{0.386} & {0.406} \\
& 336 & 0.594 & 0.541 &\textbf{0.547} & \textbf{0.515} & 0.426 & \textbf{0.433} & \textbf{0.424} & {0.437} \\
& 720 & 0.831 & 0.657 &\textbf{0.691} & \textbf{0.592} & 0.431 & 0.446 & \textbf{0.427} & \textbf{0.440} \\
\bottomrule

\end{tabular}
\label{tab:plugin}
\end{threeparttable}
\end{table*}

\renewcommand{\arraystretch}{1}

\subsubsection{Influence Analysis of Different Frequency Bands}
\begin{figure*}[!ht]
    \centering
    \begin{minipage}{0.30\textwidth}
        \centering
        \includegraphics[width=\linewidth]{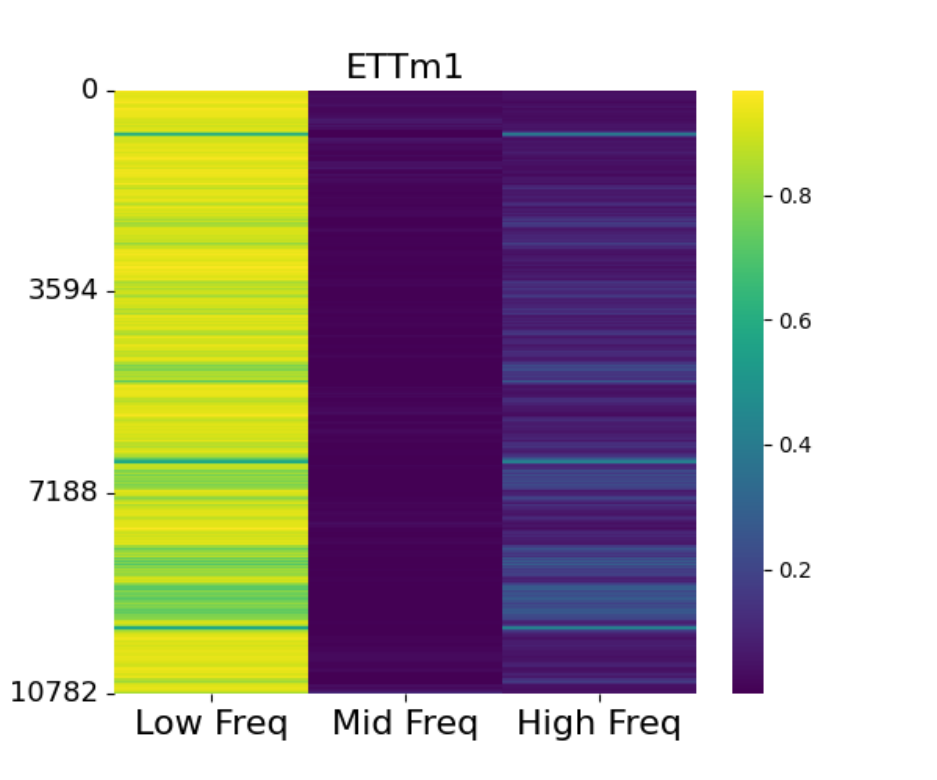}  
    \end{minipage}%
    \begin{minipage}{0.30\textwidth}
        \centering
        \includegraphics[width=\linewidth]{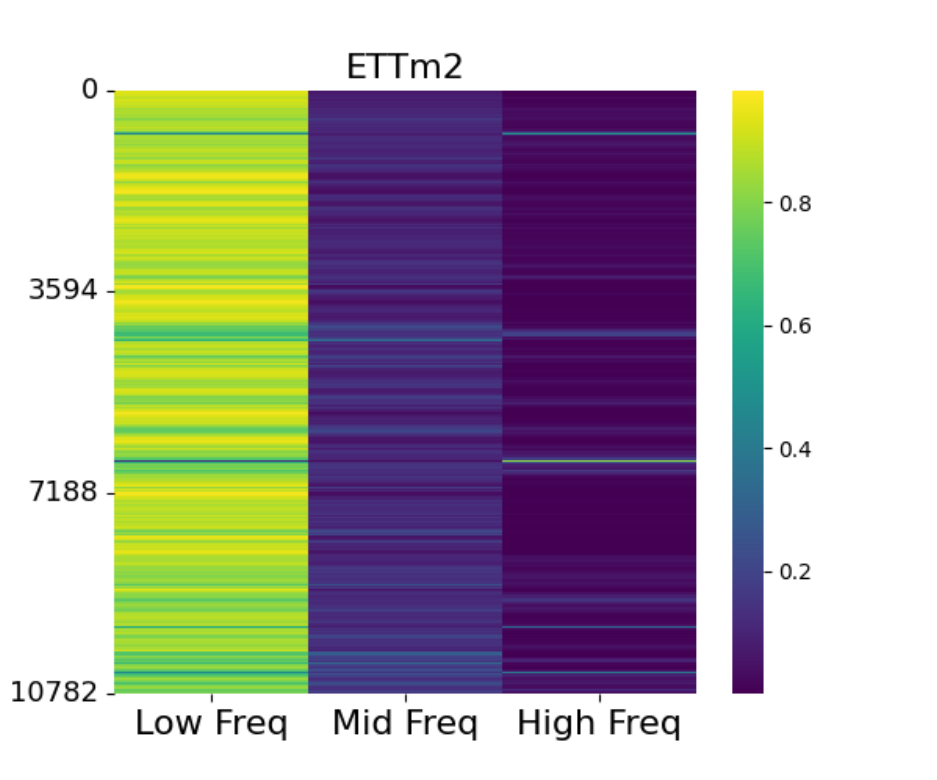}
    \end{minipage}%
    \begin{minipage}{0.30\textwidth}
        \centering
        \includegraphics[width=\linewidth]{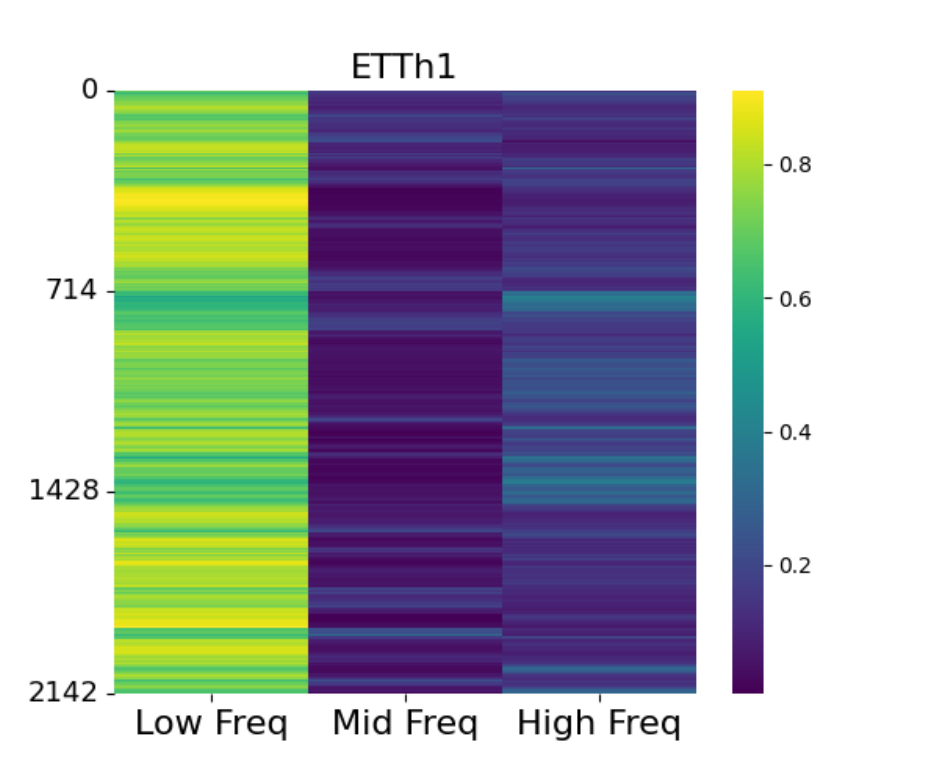}
    \end{minipage}

    \begin{minipage}{0.30\textwidth}
        \centering
        \includegraphics[width=\linewidth]{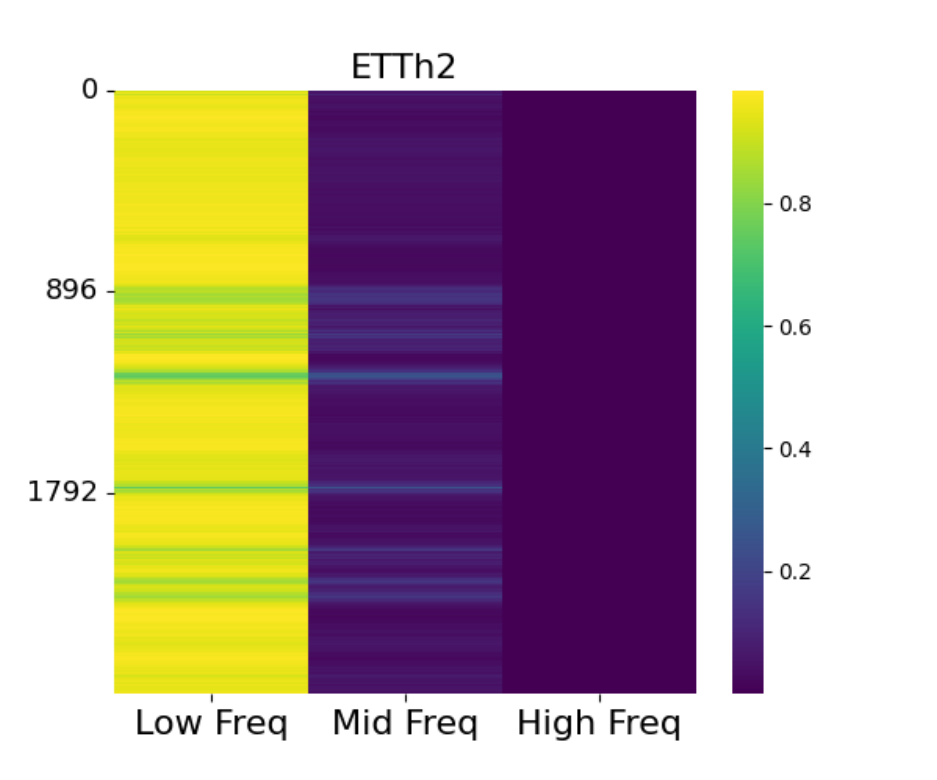}
    \end{minipage}%
    \begin{minipage}{0.30\textwidth}
        \centering
        \includegraphics[width=\linewidth]{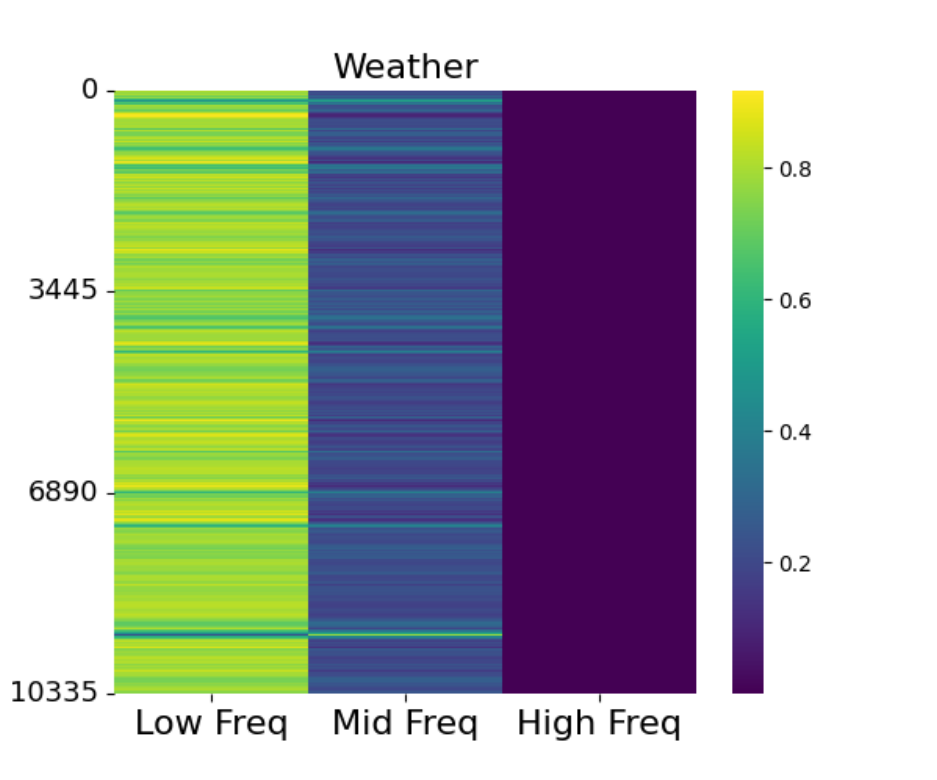}
    \end{minipage}%
    \begin{minipage}{0.30\textwidth}
        \centering
        \includegraphics[width=\linewidth]{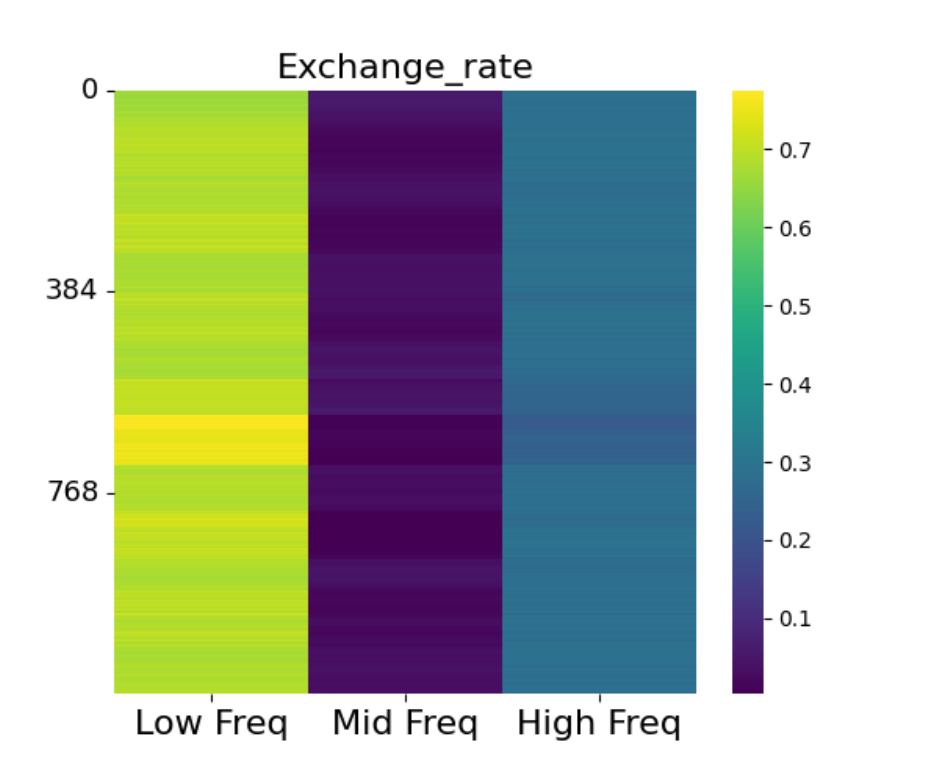}
    \end{minipage}

    \caption{Heatmap representation of gating coefficient sequences across different frequency bands. The X-axis denotes the frequency bands (experts), and the Y-axis represents the sequence indices. The color intensity in each cell corresponds to the gating coefficient, with higher values indicating a greater contribution from the respective frequency band to the prediction of the input sequence.}
    \label{fig:combined_heatmaps}
\end{figure*}
In addition to the previous ablation study, we conducted an experiment to validate the claim in Section 2.2 that excessive low-pass filtering results in significant information loss. We also systematically examined the impact of different frequency bands on time series forecasting.

In this experiment, we set the prediction length to 720 and the Mixture of Experts module to three experts, each responsible for a specific frequency band: low, mid, and high frequencies. Using six standard datasets (ETTm1, ETTm2, ETTh1, ETTh2, Weather\citep{angryk2020multivariate}, and Exchange Rate), we trained the model, extracted the gating coefficients for each expert, and visualized them in Figure \ref{fig:combined_heatmaps} as a heatmap. The X-axis shows frequency bands (experts), the Y-axis represents sequence indices, and the cell values indicate gating coefficients. This visualization highlights how different frequency bands influence each sequence, with higher values indicating greater importance of a specific band.

In the ETTm1 dataset, the heatmap shows the highest gating factor overall in the low frequency band, indicating that the main temporal pattern is concentrated in this band, while the high frequency band plays a role in the final phase of the dataset. Similarly, in the ETTm2 dataset, the low-frequency band has the greatest impact, with most of the signal strength concentrated in this band, while the mid- and high-frequency bands have less of an impact. The ETTh1 dataset exhibits a similar pattern, with both the low-frequency and high-frequency bands making some contribution, while the mid-frequency band has very little impact.The low-frequency band of the ETTh2 dataset is particularly important, which underscores the fact that the low-frequency band has more of an impact on the model predictions. In the weather dataset, both the low and mid-frequency bands have a strong contribution to the predictions. In contrast, for the Exchange\_rate dataset, the low and high frequency bands play an important role at different stages, while the low frequency band contains relatively more model information.In summary, the low-frequency band does contains essential information, but Figure \ref{fig:combined_heatmaps} shows that the mid- and high-frequency bands also hold valuable information that should not be disregarded.

The diverse distribution of coefficients across different frequency bands indicates that the contribution of various components to the prediction is highly context-dependent on the specific data. This underscores the importance of not blindly applying filters prior to experimentation. This finding further validates the argument presented in Section 2.2 and indirectly supports the effectiveness of our proposed Frequency Decomposition MoE module.

\subsubsection{Efficiency Analysis}
 Table \ref{tab:param} compares the number of trainable parameters, MACs, and inference time for various time series forecasting models on ETTh1, using a look-back window and forecasting horizon of 96. Here, $n$ denotes the number of prediction blocks in FreqMoE. 
 
 Notably, larger Transformer-based models such as PatchTST and FEDformer have millions of parameters (6.89M and 16.8M) and significantly higher MACs, at 4.26G and 10.9G respectively. Even the linear-based model TiDE has a parameter count of over 2 million.

 In contrast, FreqMoE stands out for substantially reducing parameter overhead while achieving state-of-the-art performance. It only requires  $15K$ to $70K$ parameters, a fraction of the multi-million parameter demands of Transformer-based counterparts. By leveraging a complex-domain MLP backbone and frequency decomposition, FreqMoE effectively captures essential temporal patterns with minimal computational cost.
\begin{table}[h]
\renewcommand{\arraystretch}{0.9} 
\centering
\caption{Comparison of the number of trainable parameters, MACs, and inference time of models under look-back window length $T$ = 96 and prediction length $S$ = 96 on ETT dataset.}
\begin{tabular}{l|ccc}  
\toprule
Models & Params & MACs & Infer. Time \\
\midrule
DLinear & 18.6K & 1.03M & 0.98ms \\
SCINet & 48.8K& 4.51M & 18.4ms  \\
TiDE & 2.53M &  0.28G & 31.2ms \\
PatchTST & 6.89M & 4.26G & 43.4ms \\
FEDformer & 16.8M & 10.9G & 120.5ms \\
\midrule
FreqMoE (n=1)  & \textbf{14.5K} &\textbf{ 0.79M} & 2.09ms \\ 
FreqMoE (n=3) & 43.2K & 2.38M & 3.98ms \\
FreqMoE (n=5) & 71.9K & 3.96M & 4.89ms \\
\bottomrule
\end{tabular}
\label{tab:param}
\end{table}
\section{CONCLUSION AND FUTURE WORK}
In this paper, we introduce the frequency decomposition mixture-of-experts Model (FreqMoE), a new approach for time series forecasting. By decomposing the time series data in the frequency domain and assigning different frequency bands to specialized experts, the frequency band information is fully preserved and utilized. A dynamic gating mechanism adjusts the contribution of each frequency band (expert) according to the frequency characteristics of the input data, and stackable blocks of residuals iteratively refine the predictions. Our empirical experimental results show that the frequency decomposition expert mixture model significantly outperforms existing models on various benchmark datasets. Furthermore, we have conducted ablation studies on frequency decomposition module and dynamic gating mechanism, which validate the effectiveness of our design choices. As future work, we aim to extend FreqMoE to more real-world scenarios to further evaluate its robustness. Additionally, we plan to improve its interpretability by analyzing how the gating mechanism identifies and prioritizes key frequency bands in high-dimensional datasets.
\bibliography{ref}  
\section*{CHECKLIST}
 \begin{enumerate}

 \item For all models and algorithms presented, check if you include:
 \begin{enumerate}
   \item A clear description of the mathematical setting, assumptions, algorithm, and/or model. [Yes]
   \item An analysis of the properties and complexity (time, space, sample size) of any algorithm. [Yes]
   \item (Optional) Anonymized source code, with specification of all dependencies, including external libraries. [Yes]
 \end{enumerate}

 \item For any theoretical claim, check if you include:
 \begin{enumerate}
   \item Statements of the full set of assumptions of all theoretical results. [Not Applicable]
   \item Complete proofs of all theoretical results. [Not Applicable]
   \item Clear explanations of any assumptions. [Not Applicable]     
 \end{enumerate}

 \item For all figures and tables that present empirical results, check if you include:
 \begin{enumerate}
   \item The code, data, and instructions needed to reproduce the main experimental results (either in the supplemental material or as a URL). [Yes]
   \item All the training details (e.g., data splits, hyperparameters, how they were chosen). [Yes]
         \item A clear definition of the specific measure or statistics and error bars (e.g., with respect to the random seed after running experiments multiple times). [Yes]
         \item A description of the computing infrastructure used. (e.g., type of GPUs, internal cluster, or cloud provider). [Yes]
 \end{enumerate}

 \item If you are using existing assets (e.g., code, data, models) or curating/releasing new assets, check if you include:
 \begin{enumerate}
   \item Citations of the creator If your work uses existing assets. [Yes]
   \item The license information of the assets, if applicable. [Not Applicable]
   \item New assets either in the supplemental material or as a URL, if applicable. [Not Applicable]
   \item Information about consent from data providers/curators. [Not Applicable]
   \item Discussion of sensible content if applicable, e.g., personally identifiable information or offensive content. [Not Applicable]
 \end{enumerate}

 \item If you used crowdsourcing or conducted research with human subjects, check if you include:
 \begin{enumerate}
   \item The full text of instructions given to participants and screenshots. [Not Applicable]
   \item Descriptions of potential participant risks, with links to Institutional Review Board (IRB) approvals if applicable. [Not Applicable]
   \item The estimated hourly wage paid to participants and the total amount spent on participant compensation. [Not Applicable]
\end{enumerate}

\end{enumerate}

\appendix
\thispagestyle{empty}
\onecolumn
\aistatstitle{FreqMoE: Enhancing Time Series Forecasting through Frequency Decomposition Mixture of Experts: \\
Supplementary Materials}
\section{DATASET DESCRIPTION}
\label{appendix:datset description}
In this section, we briefly describe the seven real-world datasets used in the experiments, an overview of which is given in Table \ref{tab:datasets}.
\begin{table}[ht]
    \caption{Overview of all the datasets.}
    \centering
    \begin{tabular}{l|c|c|c|c}
        \toprule
        \textbf{Dataset} & \textbf{Dim} & \textbf{Timestep} & \textbf{Prediction Length} & \textbf{Frequency} \\
        \midrule
        ETTh1     &  7&  17420& \{96,192,336,720\} & Hour \\
        ETTh2     &  7&  17420&  \{96,192,336,720\}& Hour \\
        ETTm1     &  7&  69680& \{96,192,336,720\}& 15 Minutes \\
        ETTm2     &  7&  69680& \{96,192,336,720\} & 15 Minutes \\
        Weather   &  21& 52696 &\{96,192,336,720\} & 10 Minutes \\
        Exchange  &  8&  7588& \{96,192,336,720\} &  Hour\\
        ECL       &  321&  26304 & \{96,192,336,720\} &  Hour\\
        PEMS03 & 358 & 25887 &12 & 5 Minutes \\
        PEMS04 & 307 & 16922 &12 & 5 Minutes \\
        PEMS07& 883 & 28155 &12 & 5 Minutes \\
        PEMS08 & 170 & 17786 &12 & 5 Minutes \\
        \bottomrule
    \end{tabular}
    \label{tab:datasets}
\end{table}
The ETT dataset consists of six load sequences and one oil temperature sequence, focusing on relevant time series analysis of the power system. The meteorological dataset covers 21 meteorological variables and records a wealth of meteorological information within the entire year 2020, with a data frequency of every 10 minutes, and all data are collected by the weather stations of the Max Planck Institute for Biogeochemistry. The exchange rate dataset, on the other hand, provides daily exchange rate data for eight countries over the period from 1990 to 2016, revealing trends in national currency exchange rates over time. The ECL dataset contains electricity consumption data for 321 customers, collected since January 1, 2011, covering a large number of variables and complex temporal patterns. Finally the PEMS dataset, consists of real-world traffic flow data collected from highway sensors across multiple locations in California, including four subsets, PEMS03, PEMS04, PEMS07, and PEMS08.
\section{IMPLEMENT DETAIL}
\label{appendix:implement detail}

\subsection{Experiment Detail}
\label{appendix:experiment detail}
In the experiment, we split the ETTh and PEMS dataset into training, validation, and testing sets in a ratio of 6:2:2, while the other three datasets are divided in a ratio of 7:2:1. The FreqMoE model is implemented using PyTorch\citep{paszke2019pytorch} and conducted the experiments on a computer equipped with an NVIDIA RTX 3090 GPU with 30GB of memory. Each experiment is repeat three times with different random seed, making ensure the consistency of the results.
During the model optimization process, the Adam optimizer is employed\citep{kingma2014adam}, with a default training duration of 40 epochs and a patience setting of 6 for implementing early stopping. Additionally, the learning rate is reduced to half of its previous value after each training iteration. 
\subsection{Hyperparameters}
\label{appendix:hyperparameters}
In our FreqMoE model, there are three key hyperparameters. The number of experts in the frequency decomposition Mixture of Experts module. The number of prediction blocks in the frequency domain prediction module. And the dropout rate for the two complex linear layers within each prediction block. These three important hyperparameters, along with the selection ranges for other hyperparameters, are detailed in Table \ref{tab:hyperparameters}. 
\begin{table}[ht]
    \caption{Hyperparameters used in the FreqMoE model.}
    \centering
    \setlength{\tabcolsep}{20pt}
    \begin{tabular}{l|c}
        \toprule
        \textbf{Parameter} & \textbf{Value} \\
        \midrule
        Expert num          & {2, 3, 4, 5, 6, 7, 8, 9, 10} \\
        Prediction block num & {1, 2, 3} \\ 
        Dropout rate        & {0.2, 0.3, 0.4} \\  
        Batch size          &  {8, 32, 64, 128} \\ 
        Initial learning rate & {0.001, 0.0005, 0.0001}  \\ 
        Prediction length    & {12, 96, 192, 336, 720} \\
        \bottomrule
    \end{tabular}
    \label{tab:hyperparameters}
\end{table}
In addition, during the experimental process, we observed that selecting a smaller batch size can yield a slight improvement in model performance for the ETTh dataset. Furthermore, regarding the choice of the number of experts, we recommend using 3, 5, 8, or 10 experts, as these configurations typically result in optimal performance. 
\subsection{Baseline Model}
Here is a brief description of the baseline models used in this paper.
\begin{enumerate}
    \item \textbf{iTransformer} \citep{liu2024itransformerinvertedtransformerseffective} is a Transformer-based model that embeds variable tokens at individual time points to capture correlations between multiple variables by applying attention mechanisms to inverted dimensions.
    The source code is available at: \href{https://github.com/thuml/iTransformer}{https://github.com/thuml/iTransformer}.
    \item \textbf{PatchTST} \citep{nie2023timeseriesworth64} PatchTST is a Transformer-based model that improves long-time prediction accuracy by dividing the time series into subsequence-level patches as input tokens and by using channel-independent techniques. The source code is available at: \href{https://github.com/yuqinie98/PatchTST}{https://github.com/yuqinie98/PatchTST}.
    \item \textbf{Crossformer} \citep{zhang2023crossformer} is a Transformer-based model that efficiently captures inter-temporal and inter-dimensional dependencies by embedding the time series as a two-dimensional vector array, preserving both temporal and dimensional information. The source code is available at:
    \href{https://github.com/Thinklab-SJTU/Crossformer}{https://github.com/Thinklab-SJTU/Crossformer}.
    \item \textbf{TiDE} \citep{das2024longtermforecastingtidetimeseries} is a MLP-based encoder-decoder model architecture that matches the performance of Transformer-based models on LSTF tasks. The source code is available at: \href{https://github.com/google-research/google-research/tree/master/tide}{https://github.com/google-research/google-research/tree/master/tide}.
    \item \textbf{TimesNet} \citep{wu2023timesnettemporal2dvariationmodeling} is a CNN-based model, efficiently capture complex time-varying features by converting a one-dimensional time series into a two-dimensional tensor. The source code is available at: \href{https://github.com/thuml/TimesNet}{https://github.com/thuml/TimesNet}.
    \item \textbf{DLinear} \citep{zeng2022transformerseffectivetimeseries} 
     is a MLP-based model with only one simple linear layer. The source code is available at: \href{https://github.com/cure-lab/LTSF-Linear}{https://github.com/cure-lab/LTSF-Linear}
     \item \textbf{SCINet} \citep{liu2022scinettimeseriesmodeling} is an CNN-based model that extracted different temporal features from downsampled subsequence by using multiple convolutional filters in each layer. The source code is available at: \href{https://github.com/cure-lab/SCINet}{https://github.com/cure-lab/SCINet}
     \item \textbf{FEDformer} \citep{zhou2022fedformerfrequencyenhanceddecomposed} is a Transformer-based model is a Transformer-based model proposing seasonaltrend decomposition and exploiting the sparsity of time series in the frequency domain.  The source code is available at: \href{ https://github.com/DAMO-DI-ML/ICML2022-FEDformer}{ https://github.com/DAMO-DI-ML/ICML2022-FEDformer}.
     \item \textbf{Autoformer} \citep{wu2022autoformerdecompositiontransformersautocorrelation} is a Trasnformer-based model address the lack of long-term dependencies by introducing an auto-correlation mechanism and an asymptotic decomposition architecture.
     The source code is available at: \href{https://github.com/thuml/Autoformer}{https://github.com/thuml/Autoformer}.
\end{enumerate}
\section{ADDITIONAL EXPERIMENTS RESULT}
\label{appendix:additional experiment result}
\subsection{Full Result of Comparison Expertiment}
\label{appendix:full comparison experiment}
Due to the space limitation of the main text, we place the full results of forecasting in Table \ref{tab:main_result}. The prediction lengths \( S \in \{96, 192, 336, 720\} \) and the lookback length is fixed as \( T = 96 \).
\begin{table*}[!ht]
  \vspace{-5pt}
  \renewcommand{\arraystretch}{0.8} %行间距
  \centering
  \resizebox{\textwidth}{!}{
  \scalebox{0.8}{
    \begin{threeparttable}
      \small
      \renewcommand{\multirowsetup}{\centering}
      \setlength{\tabcolsep}{1.1pt}
      \caption{Long-term multivariate forecasting results with prediction lengths \( S \in \{96, 192, 336, 720\} \) and fixed lookback length \( T = 96 \). The best forecasting results are in bold and the second-best are underlined. Lower MSE/MAE indicates more accurate predictions.The results of other models are sourced from iTransformer\citep{liu2024itransformerinvertedtransformerseffective}.}
      \begin{tabular}{c|c|cc|cc|cc|cc|cc|cc|cc|cc|cc|cc}
    \toprule
    \multicolumn{2}{c}{\multirow{1}{*}{Models}} & 
    \multicolumn{2}{c}{\rotatebox{0}{{{FreqMoE}}}} &
    \multicolumn{2}{c}{\rotatebox{0}{{iTransformer}}} &
    \multicolumn{2}{c}{\rotatebox{0}{{PatchTST}}} &
    \multicolumn{2}{c}{\rotatebox{0}{{Crossformer}}}  &
    \multicolumn{2}{c}{\rotatebox{0}{{TiDE}}} &
    \multicolumn{2}{c}{\rotatebox{0}{{{TimesNet}}}} &
    \multicolumn{2}{c}{\rotatebox{0}{{DLinear}}}&
    \multicolumn{2}{c}{\rotatebox{0}{{SCINet}}} &
    \multicolumn{2}{c}{\rotatebox{0}{{FEDformer}}} &
    \multicolumn{2}{c}{\rotatebox{0}{{Autoformer}}} \\
    \cmidrule(lr){3-4} \cmidrule(lr){5-6}\cmidrule(lr){7-8} \cmidrule(lr){9-10}\cmidrule(lr){11-12}\cmidrule(lr){13-14} \cmidrule(lr){15-16} \cmidrule(lr){17-18} \cmidrule(lr){19-20} \cmidrule(lr){21-22}
    \multicolumn{2}{c}{Metric}  & {MSE} & {MAE}  & {MSE} & {MAE}  & {MSE} & {MAE}  & {MSE} & {MAE}  & {MSE} & {MAE}  & {MSE} & {MAE} & {MSE} & {MAE} & {MSE} & {MAE} & {MSE} & {MAE} & {MSE} & {MAE}\\
    \toprule

    \multirow{5}{*}{{\rotatebox{90}{{ETTm1}}}}
    &  {96} & \textbf{{0.314}} & \textbf{{0.356}} & {{0.334}} & {{0.368}} & \underline{{0.329}} & \underline{{0.367}} & {0.404} & {0.426} & {0.364} & {0.387} &{{0.338}} &{{0.375}} &{{0.345}} &{{0.372}} & {0.418} & {0.438} &{0.379} &{0.419} &{0.505} &{0.475} \\
    & {192} & {\textbf{0.356}} & {\textbf{0.380}} & {0.377} & {0.391} & \underline{{0.367}} & \underline{{0.385}} & {0.450} & {0.451} &{0.398} & {0.404} &{{0.374}} &{{0.387}}  &{{0.380}} &{{0.389}} & {0.439} & {0.450}  &{0.426} &{0.441} &{0.553} &{0.496} \\
    & {336} & \textbf{{0.385}} & \textbf{{0.404}} & {0.426} & {0.420} & \underline{{0.399}} & \underline{{0.410}} & {0.532}  &{0.515} & {0.428} & {0.425} &{{0.410}} &{{0.411}}  &{{0.413}} &{{0.413}} & {0.490} & {0.485}  &{0.445} &{0.459} &{0.621} &{0.537} \\ 
    & {720} & \textbf{{0.446}} &\underline{ {0.445}} & {0.491} & {0.459} & \underline{{0.454}} & \textbf{{0.439}} & {0.666} & {0.589} & {0.487} & {0.461} &{{0.478}} &{{0.450}} &{{0.474}} &{{0.453}} & {0.595} & {0.550}  &{0.543} &{0.490} &{0.671} &{0.561} \\ 
    \cmidrule(lr){2-22}
    & {Avg} & \textbf{{0.375}} & \textbf{{0.396}} & {0.407} & {0.410} & \underline{{0.387}} & \underline{{0.400}} & {0.513} & {0.496} & {0.419} & {0.419} &{{0.400}} &{{0.406}}  &{{0.403}} &{{0.407}} & {0.485} & {0.481}  &{0.448} &{0.452} &{0.588} &{0.517} \\ 
    \midrule
    
    \multirow{5}{*}{\rotatebox{90}{{ETTm2}}}
    &  {96} & \textbf{{0.173}} & \underline{{0.266}} & {0.180} & {0.264} & \underline{{0.175}} & \textbf{{0.259}} & {0.287} & {0.366} & {0.207} & {0.305} &{{0.187}} &{0.267} &{0.193} &{0.292} & {0.286} & {0.377} &{0.203} &{0.287} &{0.255} &{0.339} \\ 
    & {192} & \textbf{{0.235}} & \underline{{0.310}} & {0.250} & {0.309} & \underline{{0.241}} & \textbf{{0.302}} & {0.414} & {0.492} & {0.290} & {0.364} &{{0.249}} &{{0.309}} &{0.284} &{0.362} & {0.399} & {0.445} &{0.269} &{0.328} &{0.281} &{0.340} \\ 
    & {336} & \textbf{{0.290}} & {{0.350}} & {{0.311}} & \underline{{0.348}} & \underline{{0.305}} & \textbf{{0.343}}  & {0.597} & {0.542}  & {0.377} & {0.422} &{{0.321}} &{{0.351}} &{0.369} &{0.427} & {0.637} & {0.591} &{0.325} &{0.366} &{0.339} &{0.372} \\ 
    & {720} & \textbf{{0.385}} & {{0.424}} & {{0.412}} & {{0.407}} & \underline{{0.402}} & \textbf{{0.400}} & {1.730} & {1.042} & {0.558} & {0.524} &{{0.408}} &\underline{{0.403}} &{0.554} &{0.522} & {0.960} & {0.735} &{0.421} &{0.415} &{0.433} &{0.432} \\ 
    \cmidrule(lr){2-22}
    & {Avg} & \textbf{{0.270}} &{{0.337}} & {{0.288}} & \underline{{0.332}} & \underline{{0.281}} & \textbf{{0.326}} & {0.757} & {0.610} & {0.358} & {0.404} &{{0.291}} &{{0.333}} &{0.350} &{0.401} & {0.571} & {0.537} &{0.305} &{0.349} &{0.327} &{0.371} \\ 
    \midrule

    \multirow{5}{*}{\rotatebox{90}{{{ETTh1}}}}
    &  {96} & {{\textbf{0.371}}} & \textbf{{0.388}} & {{0.386}} & {{0.405}} & {0.414} & {0.419} & {0.423} & {0.448} & {0.479}& {0.464}  &{{0.384}} &{{0.402}} & {0.386} &\underline{{0.400}} & {0.654} & {0.599} &\underline{{0.376}} &{0.419} &{0.449} &{0.459}  \\ %&{0.865} &{0.713} \\
    & {192} & \underline{{0.426}} & \textbf{{0.422}} & {0.441} & {0.436} & {0.460} & {0.445} & {0.471} & {0.474}  & {0.525} & {0.492} &{{0.436}} &\underline{{0.429}}  &{{0.437}} &{{0.432}} & {0.719} & {0.631} &\textbf{{0.420}} &{0.448} &{0.500} &{0.482} \\ %&{1.008} &{0.82} \\
    & {336} & \underline{{0.475}} & \textbf{{0.447}} & {{0.487}} & \underline{{0.458}} & {0.501} & {0.466} & {0.570} & {0.546} & {0.565} & {0.515} &{0.491} &{0.469} &{{0.481}} & {{0.459}} & {0.778} & {0.659} &\textbf{{0.459}} &{{0.465}} &{0.521} &{0.496} \\ %&{1.107} &{0.809} \\
    & {720} & \textbf{{0.488}} & \textbf{{0.459}} & {{0.503}} & {{0.491}} & \underline{{0.500}} & \underline{{0.488}} & {0.653} & {0.621} & {0.594} & {0.558} &{0.521} &{{0.500}} &{0.519} &{0.516} & {0.836} & {0.699} &{{0.506}} &{{0.507}} &{0.514} &{0.512}  \\ 
    \cmidrule(lr){2-22}
    & {Avg} & \textbf{{0.440}} & \textbf{{0.429}} & {{0.454}} & \underline{{0.447}} & {0.469} & {0.454} & {0.529} & {0.522} & {0.541} & {0.507} &{0.458} &{{0.450}} &{{0.456}} &{{0.452}} & {0.747} & {0.647} &\underline{{0.440}} &{0.460} &{0.496} &{0.487}  \\ %&{1.040} &{0.85} \\
    \midrule

    \multirow{5}{*}{\rotatebox{90}{{ETTh2}}}
    &  {96} & \textbf{{0.287}} & \textbf{{0.337}} & \underline{{0.297}} & {{0.349}} & {{0.302}} & \underline{{0.348}} & {0.745} & {0.584} &{0.400} & {0.440}  & {{0.340}} & {{0.374}} &{{0.333}} &{{0.387}} & {0.707} & {0.621}  &{0.358} &{0.397} &{0.346} &{0.388} \\
    & {192} & \textbf{{0.361}} & \textbf{{0.386}} & \underline{{0.380}} & \underline{{0.400}} &{{0.388}} & {{0.400}} & {0.877} & {0.656} & {0.528} & {0.509} & {{0.402}} & {{0.414}} &{0.477} &{0.476} & {0.860} & {0.689} &{{0.429}} &{{0.439}} &{0.456} &{0.452} \\ 
    & {336} & \textbf{{0.407}} & \textbf{{0.423}} & {{0.428}} & \underline{{0.432}} & \underline{{0.426}} & {{0.433}}& {1.043} & {0.731} & {0.643} & {0.571}  & {{0.452}} & {{0.452}} &{0.594} &{0.541} & {1.000} &{0.744} &{0.496} &{0.487} &{0.482} &{0.486}\\ 
    & {720} & \textbf{{0.414}} & \textbf{{0.438}} & \underline{{0.427}} & \underline{{0.445}} & {{0.431}} & {{0.446}} & {1.104} & {0.763} & {0.874} & {0.679} & {{0.462}} & {{0.468}} &{0.831} &{0.657} & {1.249} & {0.838} &{{0.463}} &{{0.474}} &{0.515} &{0.511} \\ 
    \cmidrule(lr){2-22}
    & {Avg} & \textbf{{0.367}} & \textbf{{0.397}} & \underline{{0.383}} & \underline{{0.407}} & {{0.387}} & {{0.407}} & {0.942} & {0.684} & {0.611} & {0.550}  &{{0.414}} &{{0.427}} &{0.559} &{0.515} & {0.954} & {0.723} &{{0.437}} &{{0.449}} &{0.450} &{0.459} \\ 
    \midrule

    \multirow{5}{*}{\rotatebox{90}{{Exchange}}}
    &  {96} & \textbf{{0.080}} & \textbf{{0.198}} & \underline{{0.086}} & {0.206} & {{0.088}} & \underline{{0.205}} & {0.256} & {0.367} & {0.094} & {0.218} & {0.107} & {0.234} & {0.088} & {0.218} & {0.267} & {0.396} & {0.148} & {0.278} & {0.197} & {0.323} \\ 
    &  {192} & \textbf{{0.170}} & \textbf{{0.293}} & {0.177} & {0.299} & \underline{{0.176}} & \underline{{0.299}} & {0.470} & {0.509} & {0.184} & {0.307} & {0.226} & {0.344} & {{0.176}} & {0.315} & {0.351} & {0.459} & {0.271} & {0.315} & {0.300} & {0.369} \\ 
    &  {336} & \textbf{0.299} & \textbf{{0.392}} & {0.331} & {0.417}& \underline{{0.301}} & \underline{{0.397}} & {1.268} & {0.883} & {0.349} & {0.431} & {0.367} & {0.448} & {{0.313}} & {0.427} & {1.324} & {0.853} & {0.460} & {0.427} & {0.509} & {0.524} \\ 
    &  {720} & \textbf{{0.826}} & \underline{{0.693}} & {0.847} & \textbf{{0.691}} & {0.901} & {0.714} & {1.767} & {1.068} & {0.852} & {0.698} & {0.964} & {0.746} & \underline{{0.839}} & {0.695} & {1.058} & {0.87} & {1.195} & {{0.695}} & {1.447} & {0.941} \\ 
    \cmidrule(lr){2-22}
    &  {Avg} & \textbf{{0.343}} & \textbf{{0.394}} & {0.360} & \underline{{0.403}} & {0.367} & {{0.404}} & {0.940} & {0.707} & {0.370} & {0.413} & {0.416} & {0.443} & \underline{{0.354}} & {0.414} & {0.750} & {0.626} & {0.519} & {0.429} & {0.613} & {0.539} \\ 
    \midrule

    \multirow{5}{*}{\rotatebox{90}{{Weather}}} 
    &  {96} & \underline{{0.168}} & \underline{{0.215}} & {0.174} & \textbf{0.214} & {0.177} & {{0.218}} & \textbf{{0.158}} & {0.230}  & {0.202} & {0.261} &{{0.172}} &{{0.220}} & {0.196} &{0.255} & {0.221} & {0.306} & {0.217} &{0.296} & {0.266} &{0.336} \\ 
    & {192} & \underline{{0.212}} & \textbf{{0.253}} & {0.221} & \underline{{0.254}} & {0.225} & {0.259} & \textbf{{0.206}} & {0.277} & {0.242} & {0.298} &{{0.219}} &{{0.261}}  & {0.237} &{0.296} & {0.261} & {0.340} & {0.276} &{0.336} & {0.307} &{0.367} \\
    & {336} & \textbf{{0.268}} & \textbf{{0.291}} & {0.278} & \underline{{0.296}} & {0.278} & {{0.297}} & \underline{{0.272}} & {0.335} & {0.287} & {0.335} &{{0.280}} &{{0.306}} & {0.283} &{0.335} & {0.309} & {0.378} & {0.339} &{0.380} & {0.359} &{0.395}\\ 
    & {720} & \textbf{{0.342}} & \textbf{{0.345}} & {0.358} & {0.349} & {{0.354}} & \underline{{0.348}} & {0.398} & {0.418} & {{0.351}} & {0.386} &{0.365} &{{0.359}} & \underline{{0.345}} &{{0.381}} & {0.377} & {0.427} & {0.403} &{0.428} & {0.419} &{0.428} \\ 
    \cmidrule(lr){2-22}
    & {Avg} & \textbf{{0.247}} & \textbf{{0.276}} & \underline{{0.258}} & \underline{{0.279}} & {{0.259}} & {{0.281}} & {0.259} & {0.315} & {0.271} & {0.320} &{{0.259}} &{{0.287}} &{0.265} &{0.317} & {0.292} & {0.363} &{0.309} &{0.360} &{0.338} &{0.382} \\ 
    \midrule

      \multirow{5}{*}{\rotatebox{90}{{ECL}}} 
    &  {96} & \underline{{0.152}} & \underline{{0.246}}& \textbf{{0.148}} & \textbf{{0.240}}  & {0.181} & {{0.270}} & {0.219} & {0.314} & {0.237} & {0.329} &{{0.168}} &{0.272} &{0.197} &{0.282} & {0.247} & {0.345} &{0.193} &{0.308} &{0.201} &{0.317}  \\ %&{0.274} &{0.368} \\
    & {192} & \underline{{0.165}} & \underline{{0.255}}& \textbf{{0.162}} & \textbf{{0.253}}  & {0.188} & {{0.274}} & {0.231} & {0.322} & {0.236} & {0.330} &{{0.184}} &{0.289} &{0.196} &{{0.285}} & {0.257} & {0.355} &{0.201} &{0.315} &{0.222} &{0.334} \\ %&{0.296} &{0.386} \\
    & {336}  & \underline{{0.181}} & \underline{{0.274}}& \textbf{{0.178}} & \textbf{{0.269}} & {0.204} & {{0.293}} & {0.246} & {0.337} & {0.249} & {0.344} &{{0.198}} &{{0.300}} &{0.209} &{{0.301}} & {0.269} & {0.369} & {0.214} & {0.329} &{0.231} & {0.338}  \\ %&{0.300} &{0.394} \\
    & {720}  & \textbf{{0.219}} & \textbf{{0.307}}& {{0.225}} & \underline{{0.317}} & {0.246} & {0.324} & {0.280} & {0.363} & {0.284} & {0.373} &\underline{{0.220}} &{{0.320}} &{0.245} &{0.333} & {0.299} & {0.390} &{0.246} &{0.355} &{0.254} &{0.361} \\ %&{0.373} &{0.439} \\
    \cmidrule(lr){2-22}
    & {Avg}  & \underline{{0.179}} &\textbf{{0.270}}& \textbf{{0.178}} & \underline{{0.270}} & {0.205} & {{0.290}} & {0.244} & {0.334} & {0.251} & {0.344} &{{0.192}} &{0.295} &{0.212} &{0.300} & {0.268} & {0.365} &{0.214} &{0.327} &{0.227} &{0.338} \\ %&{0.311} &{0.397} \\
    \midrule
     \multicolumn{2}{c|}{{{$1^{\text{st}}$ Count}}} & {\textbf{27}} & {\textbf{24}} & {\underline{4}} & {{5}} & {0} & \underline{6} & {2} & {0} & {0} & {0} & {0} & {0} & {0} & {0} & {0} & {0} & {2} & {0} & {0} & {0} \\ %& {2} & {1}\\
    \bottomrule
  \end{tabular}
  \label{tab:main_result}
    \end{threeparttable}
  }
  }
\end{table*}

As can be seen from Table \ref{tab:main_result}, our model achieves the best results in 51 out of 70 metrics. Among them, FreqMoE performs particularly well on low-channel datasets such as the ETTh (7-channel) and Exchange (8-channel) datasets, demonstrating the good fitness of our frequency domain decomposition prediction model in low-channel scenarios. In contrast, on high-channel datasets such as Weather and ECL, the performance of the model is largely weaker than Transformer-based models such as iTransformer and PatchTST, which may indicate that there is still room for improvement in the extraction of multi-dimensional dependency information by linear models. However, it is worth noting that the model still performs well in multi-dimensional long time prediction, which indicates that the frequency domain decomposition module plays an important role in major frequency band feature extraction.
\subsection{Analysis of MoE Ablation Experiment}
\label{appendix: moe ablation experiment}
\begin{figure}[!ht]
    \centering
    \begin{minipage}[b]{0.45\textwidth}
        \centering
        \includegraphics[width=\textwidth]{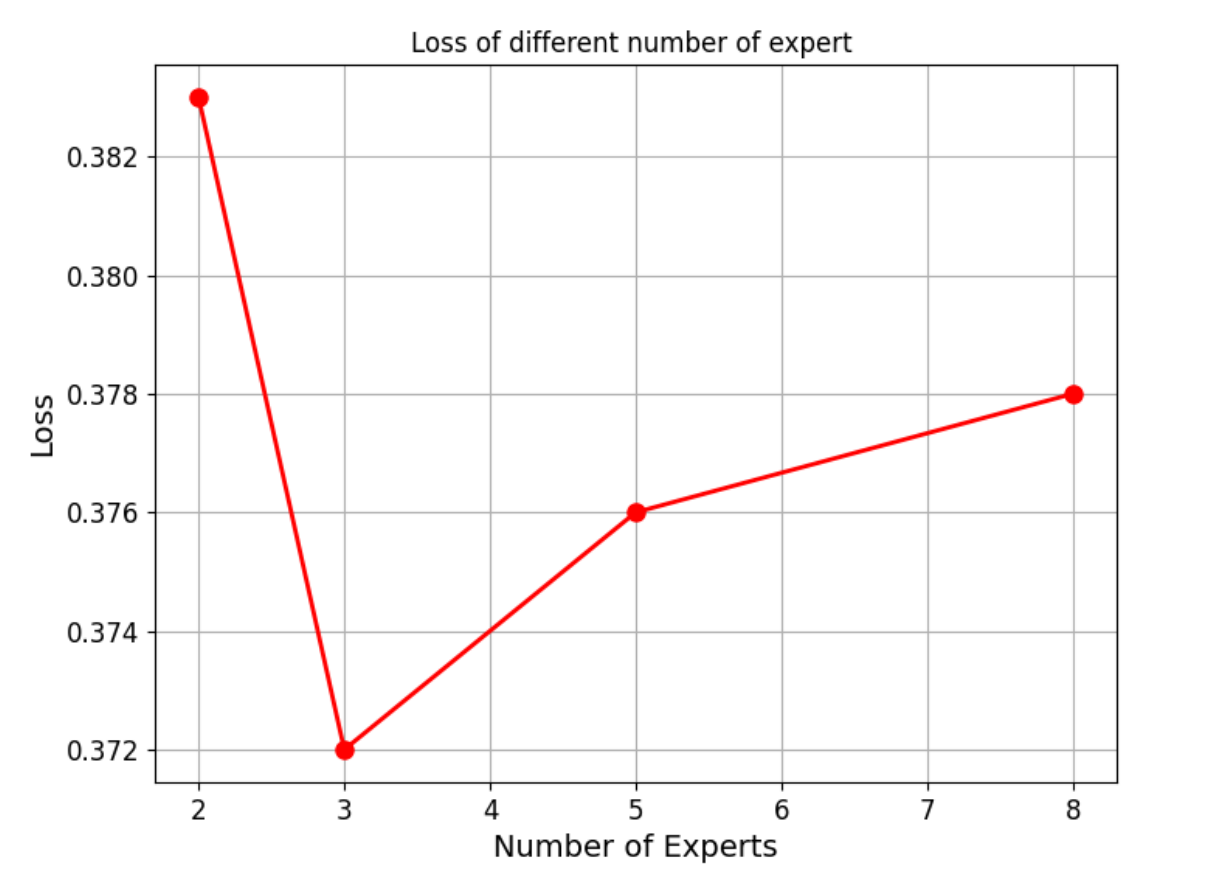}
    \end{minipage}
    \hspace{0.05\textwidth} 
    \begin{minipage}[b]{0.45\textwidth}
        \centering
        \includegraphics[width=\textwidth]{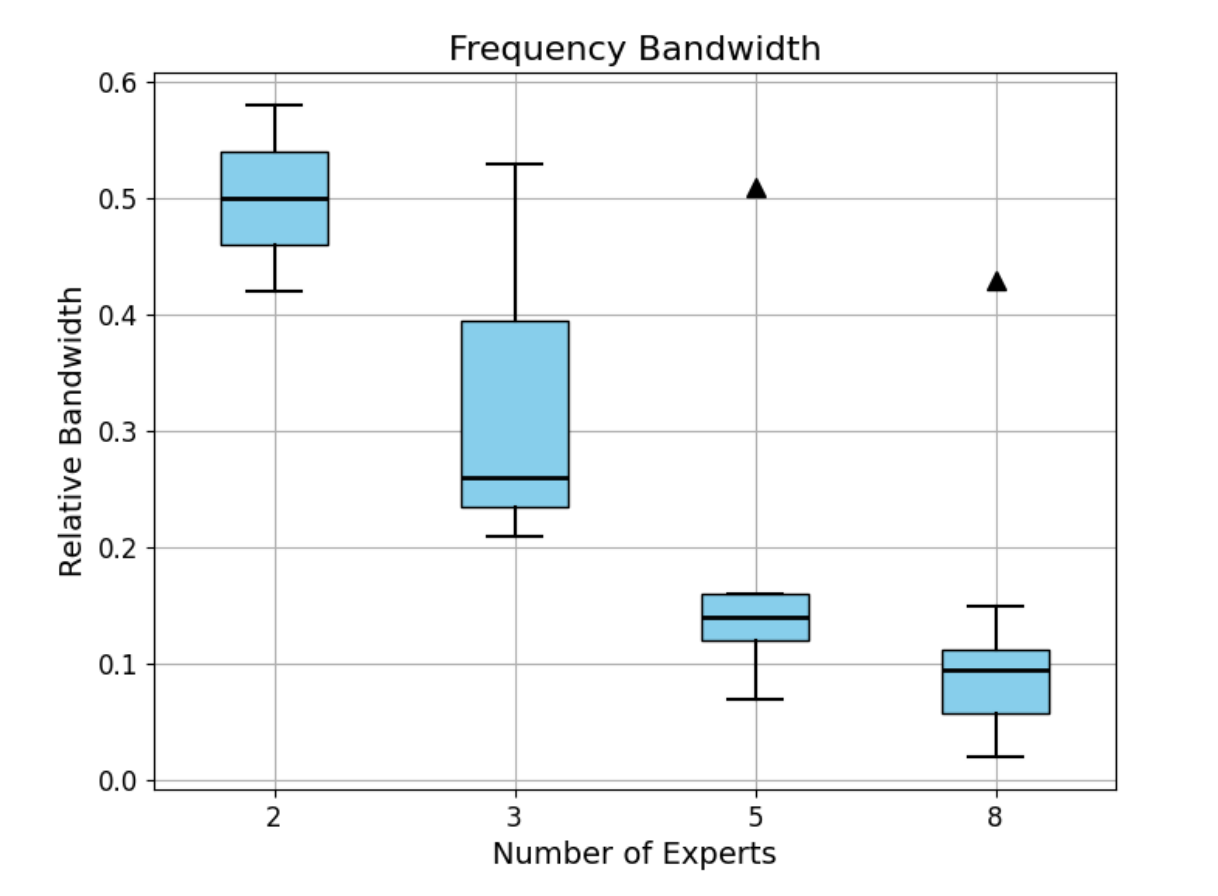}
    \end{minipage}
    \caption{The subplot on the left shows the loss for models with different numbers of experts, and the subplot on the right shows the distribution of the bandwidths of the models with different numbers of experts, where the widths are the relative widths of the bands with respect to the sequence.}
    \label{fig:moe expert}
\end{figure}

As the number of experts increases, the performance of the model not only fails to improve, but also declines. We believe that this phenomenon is mainly caused by the following two factors:
\begin{enumerate}
    \item \textbf{Frequency range fragmentation:} As the number of experts increases, the frequency range that each expert is responsible for becomes narrower, resulting in insufficient information contained within each frequency band to provide adequate feature representation.
    \item \textbf{Uneven allocation of weights in the gating network:} There is an imbalance in the allocation of weights in the gating network, with experts responsible for wider frequency ranges and containing more information being given higher weights, while other experts are given relatively lower weights. This unbalanced allocation of weights results in information in some frequency bands not being fully utilized or ignored, leading to loss of information.
\end{enumerate}
In order to verify the effect of the first factor mentioned above, we trained the models with the number of experts 2, 3, 5 and 8, and recorded the band boundaries and the final loss values for each model, by keeping the other hyperparameters of the model constant. In addition, we calculated the width of each frequency band. All the results are shown in Figure \ref{fig:moe expert}.

It can be clearly seen that as the number of experts increases, the average value of the bandwidth that each expert is responsible for decreases rapidly. When the number of experts increases to 8, except for one expert whose bandwidth is 0.43, the width of the rest of the bands is even less than 0.1. This extremely narrow bandwidth obviously leads to the amount of information contained in it being too small, thus affecting the judgment ability of the gating mechanism and weakening the performance of the model. This result validates the first factor.

\begin{figure}
    \centering
    \includegraphics[width=0.6\linewidth]{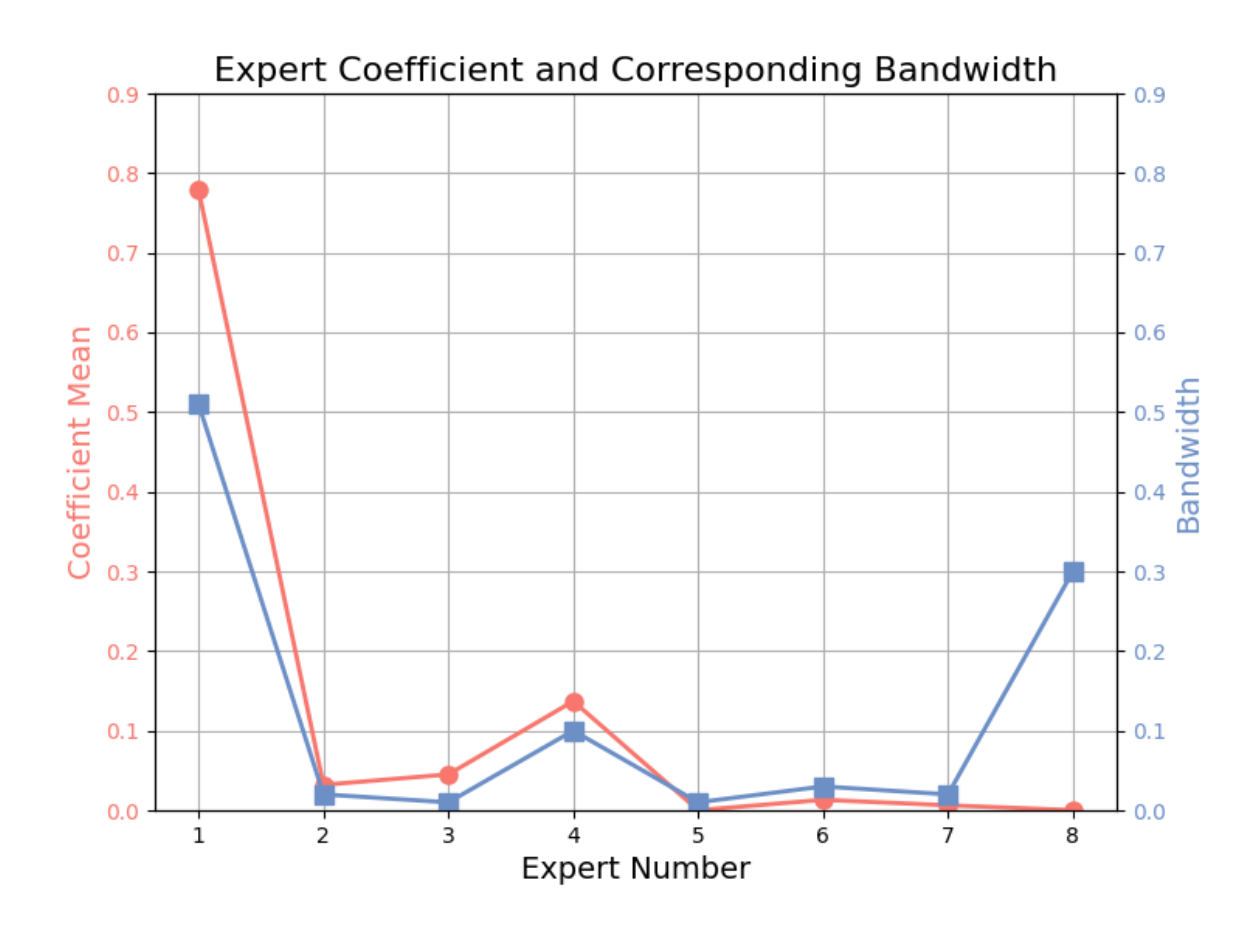}
    \caption{This figure illustrates the average coefficient assigned to each expert by the gating network, along with the bandwidth managed by each expert.}
    \label{fig:bandwidth correspond coefficient}
\end{figure}

To validate the second factor, we again trained a model containing eight experts and up-calculated the mean of the coefficients for each expert and the width of the band for which each expert was responsible on the test set. Figure \ref{fig:bandwidth correspond coefficient} illustrates the relationship between the mean value of the coefficients for each expert and their corresponding bandwidths. 

From the results, it can be seen that the experts responsible for wider frequency bands were given higher weights to some extent, while the experts with narrower frequency bands were assigned smaller weights by the gating mechanism due to their low information content. This phenomenon suggests that the gating network tends to assign higher weights to the experts responsible for the more information-rich bands, which leads to result in some of the band information being underutilized. This result validates the second factor we proposed.
\subsection{Comparison of Gating Mechanism and Fix Learnable Parameter}
\label{appendix:ablation experiment of gating mechanism}
To demonstrate the effectiveness of gating mechanism in frequency decomposition module, we conduct the ablation experiment of it, the result are shown in Table \ref{tab:gate} as follow.
\begin{table}[!ht]
    \centering
    \setlength{\tabcolsep}{18pt} % 增加列的间距以扩展表格宽度
    \caption{Performance comparison of models using gating mechanism and using fixed learnable parameter.The best results are in bold.}
    \begin{tabular}{c c c c c c}
        \toprule
        \multicolumn{1}{c}{\textbf{Dataset}} & \multicolumn{1}{c}{\textbf{Model}} & \multicolumn{2}{c}{\textbf{Gate Mechanism}} & \multicolumn{2}{c}{\textbf{Fixed Parameter}} \\
        \cmidrule(lr){3-4} \cmidrule(lr){5-6}
        \multicolumn{1}{c}{\textbf{}} & \multicolumn{1}{c}{\textbf{Metric}} & \multicolumn{1}{c}{\textbf{MSE}} & \multicolumn{1}{c}{\textbf{MAE}} & \multicolumn{1}{c}{\textbf{MSE}} & \multicolumn{1}{c}{\textbf{MAE}} \\
        \midrule
        \textbf{ETTH1} & 96  & \textbf{0.371} & \textbf{0.388} & 0.375 & 0.398 \\
        ~              & 192 & \textbf{0.426} & \textbf{0.422} & 0.428 & 0.431 \\
        ~              & 336 & \textbf{0.475} & \textbf{0.447} & 0.483 & 0.465 \\
        ~              & 720 & \textbf{0.488} & 0.459 & 0.540 & \textbf{0.420} \\
        \midrule
        \textbf{ETTH2} & 96  & \textbf{0.287} & \textbf{0.337} & 0.294 & 0.364 \\
        ~              & 192 & \textbf{0.361} & \textbf{0.386} & 0.369 & 0.424 \\
        ~              & 336 & \textbf{0.407} & \textbf{0.430} & 0.429 & 0.466 \\
        ~              & 720 & \textbf{0.414} & \textbf{0.438} & 0.447 & 0.488 \\
        \midrule
        \textbf{ETTM1} & 96  & \textbf{0.314} & \textbf{0.356} & 0.314 & 0.357 \\
        ~              & 192 & \textbf{0.356} & \textbf{0.380} & 0.357 & 0.384 \\
        ~              & 336 & \textbf{0.385} & \textbf{0.404} & 0.387 & 0.405 \\
        ~              & 720 & \textbf{0.446} & \textbf{0.442} & 0.455 & 0.450 \\
        \midrule
        \textbf{ETTM2} & 96  & \textbf{0.173} & \textbf{0.265} & 0.176 & 0.269 \\
        ~              & 192 & \textbf{0.235} & \textbf{0.310} & 0.239 & 0.318 \\
        ~              & 336 & \textbf{0.290} & \textbf{0.350} & 0.301 & 0.369 \\
        ~              & 720 & \textbf{0.385} & \textbf{0.424} & 0.406 & 0.448 \\
        \bottomrule
    \end{tabular}
    \label{tab:gate}
\end{table}

According to Table \ref{tab:gate}, the performance of the model using the gating mechanism is much better than that using the fixed parameters, which shows that the gating mechanism contributes greatly to the enhancement of the model generalization ability as well as the extraction of the key frequency modes.

\subsection{Robustness Experiment}
\label{appendix:Robustness experiment}
In addition, we performed robustness tests on all datasets used in our main text comparison experiments. We used three different random seeds (2020, 2021,2022) to ensure that the model parameters were initialized differently each time. Table \ref{tab:robustness} shows the mean and standard derivation of the MSE and MAE from multiple experiments, according to the table we can see that the variance of all three experiments is small, indicating that the random initialization has a minimal effect on the model, demonstrating the robustness of our model.
\begin{table*}[!ht]
  \vspace{-5pt}
  \renewcommand{\arraystretch}{1.0} % 行间距
  \centering
  \resizebox{\textwidth}{!}{
    \begin{threeparttable}
      \small
      \setlength{\tabcolsep}{12pt}
      \caption{Multivariate forecasting results with different random seeds(2020,2021,2022) in FreqMoE with lookback length T = 96. }
      \begin{tabular}{c|c|cc|cc|cc|cc}
        \toprule
        \multicolumn{2}{c}{\multirow{1}{*}{Models}} & 
        \multicolumn{2}{c}{\rotatebox{0}{{FreqMoE}}} &
        \multicolumn{2}{c}{\rotatebox{0}{{iTransformer}}} &
        \multicolumn{2}{c}{\rotatebox{0}{{PatchTST}}} &
        \multicolumn{2}{c}{\rotatebox{0}{{DLinear}}} \\
        \cmidrule(lr){3-4} \cmidrule(lr){5-6} \cmidrule(lr){7-8} \cmidrule(lr){9-10}
        \multicolumn{2}{c}{Metric}  & {MSE} & {MAE}  & {MSE} & {MAE}  & {MSE} & {MAE}  & {MSE} & {MAE} \\
        \midrule
    
        \multirow{5}{*}{\rotatebox{90}{ETTm1}}
        & {96} & \textbf{0.314}\ensuremath{\pm}0.002 & \textbf{0.356}\ensuremath{\pm}0.001 & 0.334 & 0.368 & \underline{0.329} & \underline{0.367} & 0.505 & 0.475 \\
        & {192} & \textbf{0.356}\ensuremath{\pm}0.004 & \textbf{0.380}\ensuremath{\pm}0.003 & 0.377 & 0.391 & \underline{0.367} & \underline{0.385} & 0.553 & 0.496 \\
        & {336} & \textbf{0.385}\ensuremath{\pm}0.004 & \textbf{0.404}\ensuremath{\pm}0.001 & 0.426 & 0.420 & \underline{0.399} & \underline{0.410} & 0.621 & 0.537 \\ 
        & {720} & \textbf{0.446}\ensuremath{\pm}0.004 & \underline{0.445}\ensuremath{\pm}0.002 & 0.491 & 0.459 & \underline{0.454} & \textbf{0.439} & 0.671 & 0.561 \\ 
        \cmidrule(lr){2-10}
        & {Avg} & \textbf{0.375}\ensuremath{\pm}0.003 & \textbf{0.396}\ensuremath{\pm}0.001 & 0.407 & 0.410 & \underline{0.387} & \underline{0.400} & 0.588 & 0.517 \\ 
        \midrule
        
        \multirow{5}{*}{\rotatebox{90}{ETTm2}}
        & {96} & \textbf{0.173}$\pm0.001$ & \underline{0.266}$\pm0.001$ & 0.180 & 0.264 & \underline{0.175} & \textbf{0.259} & 0.339 & 0.388 \\ 
        & {192} & \textbf{0.235}$\pm0.001$ & \underline{0.310}$\pm0.001$ & 0.250 & 0.309 & \underline{0.241} & \textbf{0.302} & 0.340 & 0.371 \\ 
        & {336} & \textbf{0.290}$\pm0.001$ & 0.350$\pm0.001$ & 0.311 & \underline{0.348} & \underline{0.305} & \textbf{0.343} & 0.372 & 0.371 \\ 
        & {720} & \textbf{0.385}$\pm0.002$ & 0.424$\pm0.002$ & 0.412 & 0.407 & \underline{0.402} & \textbf{0.400} & 0.459 & 0.459 \\ 
        \cmidrule(lr){2-10}
        & {Avg} & \textbf{0.270}$\pm0.001$ & 0.337$\pm0.001$ & 0.288 & \underline{0.332} & \underline{0.281} & \textbf{0.326} & 0.450 & 0.371 \\ 
        \midrule
    
        \multirow{5}{*}{\rotatebox{90}{ETTh1}}
        & {96} & \textbf{0.371}$\pm0.001$ & \textbf{0.388}$\pm0.001$ & 0.386 & 0.405 & 0.414 & 0.419 & 0.449 & 0.459 \\
        & {192} & \underline{0.426}$\pm0.001$ & \textbf{0.422}$\pm0.001$ & 0.441 & 0.436 & 0.460 & 0.445 & 0.500 & 0.482 \\
        & {336} & \underline{0.475}$\pm0.002$ & \textbf{0.447}$\pm0.003$ & 0.487 & \underline{0.458} & 0.501 & 0.466 & 0.521 & 0.496 \\
        & {720} & \textbf{0.488}$\pm0.010$ & \textbf{0.459}$\pm0.008$ & 0.503 & 0.491 & \underline{0.500} & \underline{0.488} & 0.514 & 0.512 \\ 
        \cmidrule(lr){2-10}
        & {Avg} & \textbf{0.440}$\pm0.003$ & \textbf{0.429}$\pm0.003$ & 0.454 & \underline{0.447} & 0.469 & 0.454 & 0.496 & 0.487 \\ 
        \midrule
    
        \multirow{5}{*}{\rotatebox{90}{ETTh2}}
        & {96} & \textbf{0.287}$\pm0.001$ & \textbf{0.337}$\pm0.001$ & \underline{0.297} & 0.349 & 0.302 & \underline{0.348} & 0.388 & 0.459 \\
        & {192} & \textbf{0.361}$\pm0.001$ & \textbf{0.386}$\pm0.001$ & \underline{0.380} & \underline{0.400} & 0.388 & 0.400 & 0.452 & 0.459 \\
        & {336} & \textbf{0.407}$\pm0.001$ & \textbf{0.423}$\pm0.003$ & 0.428 & \underline{0.432} & \underline{0.426} & 0.433 & 0.482 & 0.459 \\ 
        & {720} & \textbf{0.414}$\pm0.005$ & \textbf{0.438}$\pm0.006$ & \underline{0.427} & \underline{0.445} & 0.431 & 0.446 & 0.450 & 0.459 \\ 
        \cmidrule(lr){2-10}
        & {Avg} & \textbf{0.367}$\pm0.002$ & \textbf{0.397}$\pm0.003$ & 0.383 & \underline{0.407} & 0.387 & 0.407 & 0.450 & 0.459 \\ 
        \midrule
    
        \multirow{5}{*}{\rotatebox{90}{Exchange}}
        & {96} & \textbf{0.080}$\pm0.002$ & \textbf{0.198}$\pm0.001$ & \underline{0.086} & 0.206 & 0.088 & \underline{0.205} & 0.323 & 0.539 \\ 
        & {192} & \textbf{0.170}$\pm0.005$ & \textbf{0.293}$\pm0.003$ & 0.177 & 0.299 & \underline{0.176} & \underline{0.299} & 0.369 & 0.539 \\ 
        & {336} & \textbf{0.299}$\pm0.005$ & \textbf{0.392}$\pm0.012$ & 0.331 & 0.417 & \underline{0.301} & \underline{0.397} & 0.524 & 0.539 \\ 
        & {720} & \textbf{0.826}$\pm0.005$ & \underline{0.693}$\pm0.007$ & 0.847 & \textbf{0.691} & 0.901 & 0.714 & 0.941 & 0.539 \\ 
        \cmidrule(lr){2-10}
        & {Avg} & \textbf{0.343}$\pm0.004$ & \textbf{0.394}$\pm0.005$ & 0.360 & \underline{0.403} & 0.367 & 0.404 & 0.539 & 0.539 \\ 
        \midrule
    
        \multirow{5}{*}{\rotatebox{90}{Weather}} 
        & {96} & \underline{0.168}$\pm0.011$ & \underline{0.215}$\pm0.006$ & 0.174 & \textbf{0.214} & 0.177 & 0.218 & 0.336 & 0.428 \\
        & {192} & \underline{0.212}$\pm0.005$ & \textbf{0.253}$\pm0.003$ & 0.221 & \underline{0.254} & 0.225 & 0.259 & 0.382 & 0.428 \\
        & {336} & \textbf{0.268}$\pm0.003$ & \textbf{0.291}$\pm0.002$ & 0.278 & \underline{0.296} & 0.278 & 0.297 & 0.395 & 0.428 \\ 
        & {720} & \textbf{0.342}$\pm0.003$ & \textbf{0.345}$\pm0.005$ & 0.358 & 0.349 & 0.354 & \underline{0.348} & 0.428 & 0.428 \\ 
        \cmidrule(lr){2-10}
        & {Avg} & \textbf{0.247}$\pm0.005$ & \textbf{0.276}$\pm0.003$ & \underline{0.258}& \underline{0.279} & 0.259 & 0.281 & 0.382 & 0.428 \\ 
        \midrule
    
          \multirow{5}{*}{\rotatebox{90}{ECL}} 
        & {96} & \underline{0.152}$\pm0.002$ & \underline{0.246}$\pm0.001$ & \textbf{0.148} & \textbf{0.240} & 0.181 & 0.270 & 0.317 & 0.338 \\ 
        & {192} & \underline{0.165}$\pm0.001$ & \underline{0.255}$\pm0.001$ & \textbf{0.162} & \textbf{0.253} & 0.188 & 0.274 & 0.334 & 0.338 \\ 
        & {336} & \underline{0.181}$\pm0.001$ & \underline{0.274}$\pm0.001$ & \textbf{0.178} & \textbf{0.269} & 0.204 & 0.293 & 0.338 & 0.338 \\ 
        & {720} & \textbf{0.219}$\pm0.003$& \textbf{0.307}$\pm0.002$ & 0.225 & \underline{0.317} & 0.246 & 0.324 & 0.361 & 0.338 \\ 
        \cmidrule(lr){2-10}
        & {Avg} & \underline{0.179}$\pm0.001$ & \textbf{0.270} $\pm0.001$& \textbf{0.178} & \underline{0.270} & 0.205 & 0.290 & 0.338 & 0.338 \\ 
        \bottomrule
      \end{tabular}
      \label{tab:robustness}
    \end{threeparttable}
    }
\end{table*}
\subsection{Impact of the Number of Prediction Blocks}
In order to explore the effect of the number of prediction blocks connected via residuals on prediction accuracy, we designed and implemented model comparison experiments with different numbers of prediction blocks. While keeping the number of experts in the frequency domain decomposition module unchanged, we tested the performance of the model with the number of prediction blocks of 1, 2 and 3 on the Weather and Exchange datasets, respectively. The experimental results are shown in Table \ref{tab:predction block num}.
\begin{table}[!ht]
    \centering
    \setlength{\tabcolsep}{10pt} % Adjusted for additional columns
    \caption{Performance comparison of models using gating mechanism, fixed learnable parameter, and three prediction blocks. The best results are in \textbf{bold}.}
    \begin{tabular}{c c c c c c c c}
        \toprule
        \multicolumn{1}{c}{\textbf{Dataset}} & 
        \multicolumn{1}{c}{\textbf{Model Metric}} & 
        \multicolumn{2}{c}{\textbf{1 Prediction Block}} & 
        \multicolumn{2}{c}{\textbf{2 Prediction Block}} & 
        \multicolumn{2}{c}{\textbf{3 Prediction Block}} \\
        \cmidrule(lr){3-4} \cmidrule(lr){5-6} \cmidrule(lr){7-8}
        \multicolumn{1}{c}{} & 
        \multicolumn{1}{c}{\textbf{Metric}} & 
        \multicolumn{1}{c}{\textbf{MSE}} & 
        \multicolumn{1}{c}{\textbf{MAE}} & 
        \multicolumn{1}{c}{\textbf{MSE}} & 
        \multicolumn{1}{c}{\textbf{MAE}} & 
        \multicolumn{1}{c}{\textbf{MSE}} & 
        \multicolumn{1}{c}{\textbf{MAE}} \\
        \midrule
        \textbf{Exchange} & 96  & 0.081 & 0.198 & 0.083 & 0.200 & \textbf{0.080} & \textbf{0.198} \\
        ~               & 192 & 0.174 & 0.296 & 0.185 & 0.304 & \textbf{0.170} & \textbf{0.293} \\
        ~               & 336 & 0.329 & 0.410 & 0.309 & 0.408 &  \textbf{0.301} & \textbf{0.396} \\
        ~               & 720 & 0.860 & 0.706 & 0.846 & 0.704 & \textbf{0.828} & \textbf{0.697} \\
        \midrule
        \textbf{Weather} & 96  & 0.177 & 0.227 & \textbf{0.168} & \textbf{0.217} & 0.169 & 0.220 \\
        ~               & 192 & 0.229 & 0.263 & 0.218 & 0.257 & \textbf{0.214} &\textbf{0.255} \\
        ~               & 336 & 0.280 & 0.308 & 0.269 & 0.295 & \textbf{0.267} & \textbf{0.291} \\
        ~               & 720 & 0.361 & 0.359 & 0.350 & 0.351 & \textbf{0.345} & \textbf{0.349} \\
        \bottomrule
    \end{tabular}
    \label{tab:predction block num}
\end{table}

According to Table \ref{tab:predction block num}, it can be seen that a deeper network structure, increasing the number of prediction blocks, especially over a longer prediction horizon, can significantly improve the performance of the model. This is because through residual connectivity, subsequent prediction blocks are able to focus on frequency modes that were not captured by the previous prediction block, thus making fuller use of information from all frequency bands to iteratively optimize the prediction results.
\subsection{Short-Term Forecast on PEMS Dataset}
\label{appendix: short term forecast}
We evaluate FreqMoE on the PEMS dataset to test its short-term forecasting ability. The PEMS dataset is a real-world traffic flow dataset collected from highway sensors across multiple locations. It includes four subsets: PEMS03, PEMS04, PEMS07, and PEMS08, each consisting of multivariate time series data. The dataset exhibits complex frequency and multi-scale signal characteristics.

Unlike other long-term forecasting datasets, the PEMS dataset emphasizes short-term dependencies and rapidly changing dynamics. These characteristics necessitate modifications to our experimental setup to better align with the nature of short-term forecasting. Thus we conduct experiments with an input length of 96 and prediction horizon of 12,. For evaluation, we compare FreqMoE against state-of-the-art models including TimeMixer and PatchTST. We use three standard metrics \textbf{MAE}, \textbf{MAPE}, \textbf{RMSE}, following the evaluation standards used in TimeMixer\citep{wang2024timemixerdecomposablemultiscalemixing}, to access forecasting accuracy.

Table \ref{tab:pems_results} presents the forecasting results across the four PEMS datasets. From the table, it can be seen that FreqMoE consistently achieves state-of-the-art performance across most metrics, demonstrating its ability to effectively capture short-term dependencies and adapt to sudden fluctuations in traffic flow. In contrast, PatchTST performs notably worse on the PEMS dataset, aligning with observations from iTransformer. The performance may stem from the highly fluctuating nature of the dataset, where the patching mechanism of PatchTST struggles to maintain focus on localized patterns, leading to difficulties in adapting to rapid variations in traffic dynamics\citep{liu2024itransformerinvertedtransformerseffective}. Although FreqMoE achieves superior performance on most metrics, it is outperformed by TimeMixer on the PEMS07 dataset, which, with 883 channels, exhibits significantly higher dimensionality than the other subsets. This may be due to the increased complexity of frequency band interactions in high-dimensional data, making it more challenging to effectively extract key frequency components.

\begin{table}[!htp]
    \centering
    \caption{Short-Term forecasting results on the PEMS dataset. The best results are in \textbf{bold}. The lookback window length $T =  96$, and the prediction length $S = 12$. Lower MAE, MAPE, or RMSE indicates better performance.}
    \label{tab:pems_results}
    \renewcommand{\arraystretch}{1.2}
    \resizebox{\textwidth}{!}{  % 适配页面宽度
    \begin{tabular}{lccc|ccc|ccc|ccc}
        \toprule
        Dataset & \multicolumn{3}{c}{PEMS03} & \multicolumn{3}{c}{PEMS04} & \multicolumn{3}{c}{PEMS07} & \multicolumn{3}{c}{PEMS08} \\
        \cmidrule(lr){2-4} \cmidrule(lr){5-7} \cmidrule(lr){8-10} \cmidrule(lr){11-13}
        Metric & MAE & MAPE & RMSE & MAE & MAPE & RMSE & MAE & MAPE & RMSE & MAE & MAPE & RMSE \\
        \midrule
        FreqMoE  & \textbf{15.10} & \textbf{16.01} & \textbf{24.12} & \textbf{19.53} & \textbf{14.20} & \textbf{31.52} & \textbf{22.49} & 9.37 & 37.66 & \textbf{16.94} & \textbf{10.55} & 26.41 \\
        TimeMixer & 15.86 & 16.51 & 25.20 & 21.06 & 14.37 & 32.11 & 22.79 & \textbf{9.32} & \textbf{33.78} & 17.73 & 11.43 & \textbf{26.16} \\
        PatchTST  & 18.06 & 17.54 & 28.36 & 26.28 & 17.65 & 39.74 & 27.52 & 12.84 & 40.07 & 19.03 & 12.47 & 29.22 \\
        \bottomrule
    \end{tabular}
    }
\end{table}

\subsection{Synthetic Dataset Experiment}
\label{appendix:synthetic dataset experiment}
\subsubsection{Synthetic Dataset Construction}

To systematically evaluate the effectiveness of the proposed frequency decomposition mechanism, we construct a synthetic dataset designed to simulate signals with distinct frequency characteristics. The dataset consists of alternating segments where either low-frequency components dominate with high-frequency noise or high-frequency components dominate with low-frequency noise. 

The total length of dataset is set to \( T = 10,000 \) time steps, with each segment spanning \( L = 500 \) steps before switching between different frequency compositions. The number of segments $N$ is 20. For each segment \( i \), where \( i \in \{0,1,2,\dots,N-1\} \), the generated signal is defined as:

\begin{equation}
x_i(t) = A_{\text{main}} \sin(2\pi f_{\text{main}, i} t) + A_{\text{noise}} \sin(2\pi f_{\text{noise}, i} t),
\end{equation}

where the primary and noise frequency components are assigned as:

\begin{equation}
f_{\text{main}, i} =
\begin{cases} 
f_{\text{low}}, & \text{if } i \text{ is even (low-frequency dominant)} \\
f_{\text{high}}, & \text{if } i \text{ is odd (high-frequency dominant)}
\end{cases},
\end{equation}

We set the low and high-frequency components as \( f_{\text{low}} = 0.05 \) Hz and \( f_{\text{high}} = 0.4 \) Hz, respectively. The amplitude of the primary signal is fixed at \( A_{\text{main}} = 1.0 \), while the amplitude of the noise component is set to \( A_{\text{noise}} = 0.3 \).

To introduce real-world variability, we further add an independent Gaussian noise component,

\begin{equation}
\epsilon(t) \sim \mathcal{N}(0, \sigma^2), \quad \sigma = 0.1,
\end{equation}

\begin{figure}[h]
    \centering
    \captionsetup[subfigure]{font=scriptsize}
    \subfloat[Segment of the synthetic dataset]{%
        \includegraphics[width=0.3\textwidth]{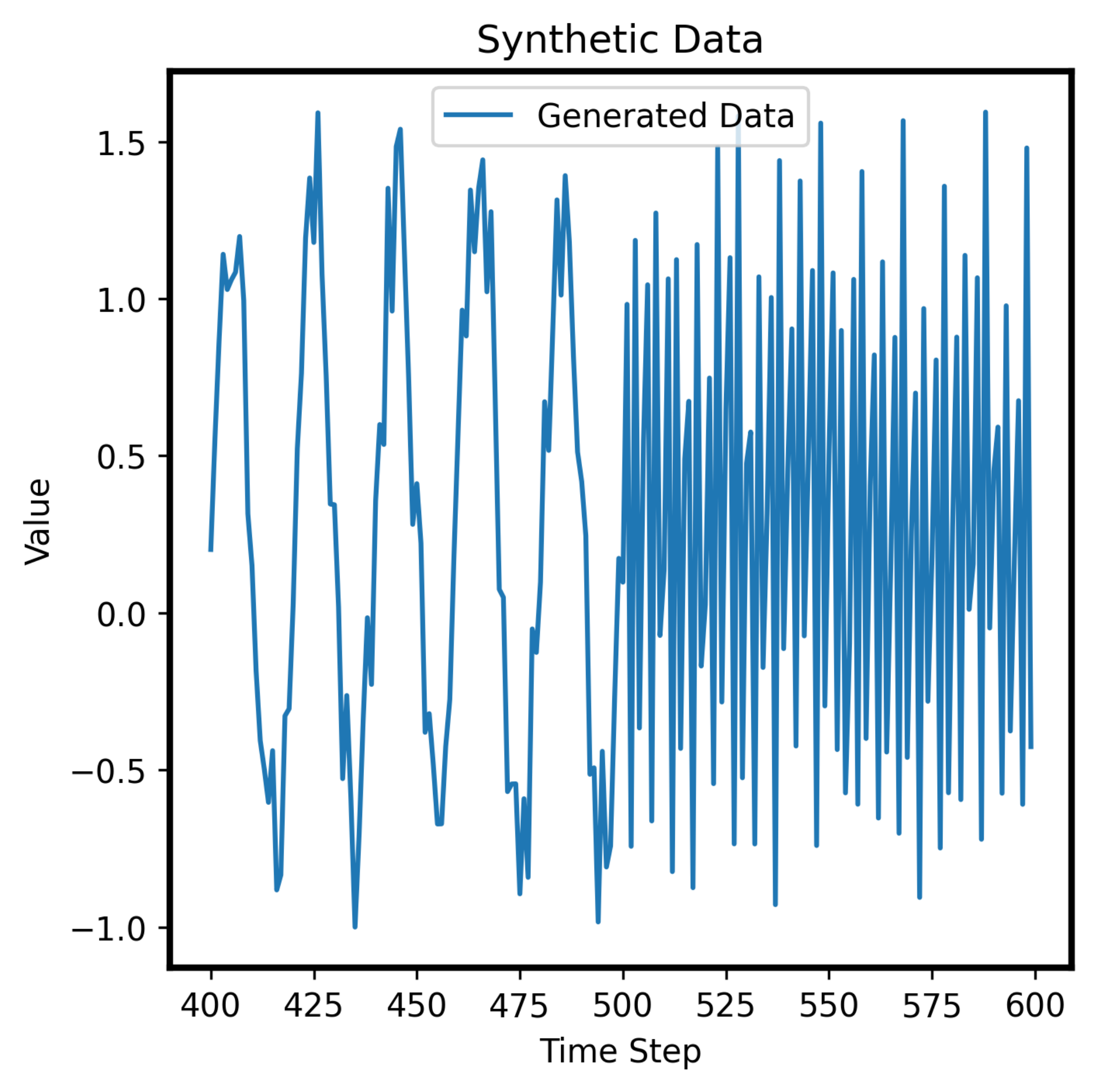}
        \label{fig:sub1}
    }
    \subfloat[  The variation of expert weights\\ over time]{%
        \includegraphics[width=0.3\textwidth]{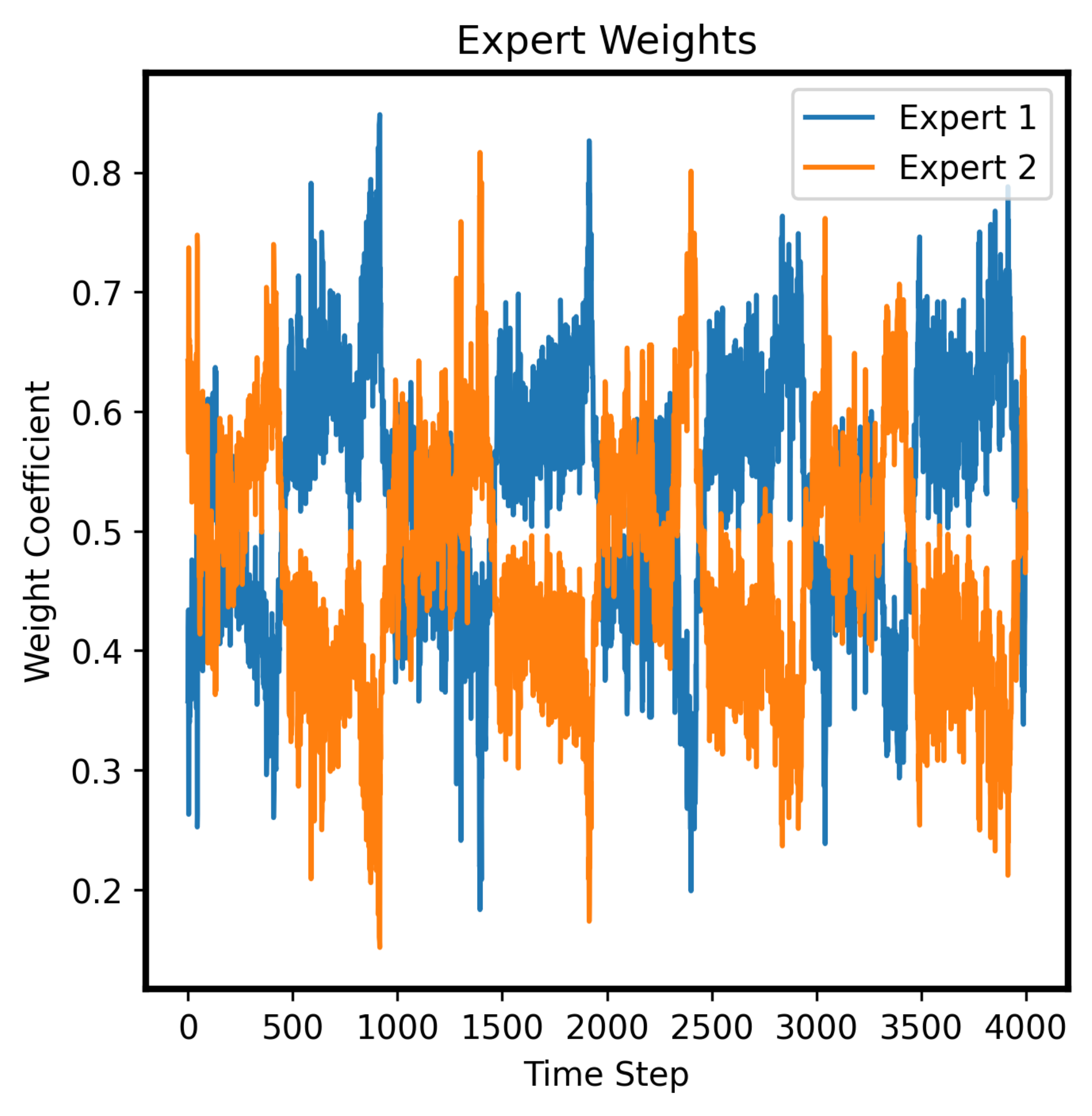}
        \label{fig:sub2}
    }
    \subfloat[ The spectrogram of the synthetic \\dataset over time step]{%
        \includegraphics[width=0.3\textwidth]{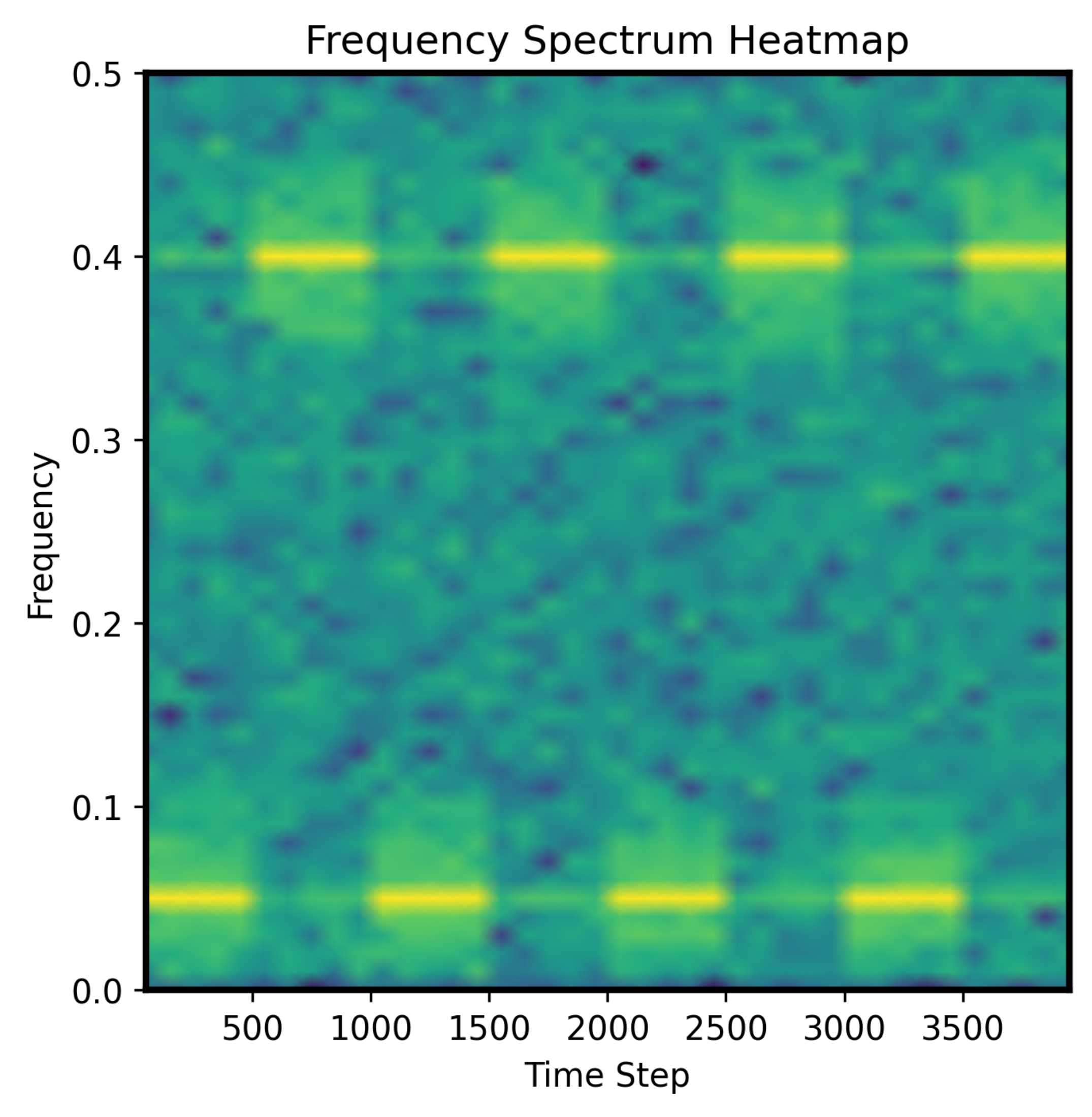}
        \label{fig:sub3}
    }
    \caption{Visualization of the synthetic dataset and expert weight. In heatmaps, positive values are shown with bright colors, while negative values are represented by darker shades.}
    \label{fig synthetic dataset}
\end{figure}

Figure \ref{fig:sub1} illustrates a sample of the Synthetic Dataset.

\subsubsection{Result and Analysis of Synthetic Dataset}
We conduct experiment with an input length $T = 96$ and prediction length $S = 96$, while fixing the number of experts to 2. As the result, Figure~\ref{fig:sub2} illustrates the evolution of expert weights over time, showing how different experts are activated in response to changing frequency characteristics. Figure~\ref{fig:sub3} presents the spectrogram of the synthetic dataset, where brighter regions indicate stronger intensity at specific frequencies. By comparing the two figures, we observe that when low-frequency components dominate, Expert 2 receives higher gating scores, while in high-frequency dominant regions, Expert 1 gains more weight. This confirms that FreqMoE effectively adapts to frequency by emphasizing dominate frequency bands while suppressing noise through dynamically adjusting the weight of different experts.

\section{SOURCE CODE}
\textbf{The code associated with this paper can be found at: }\href{https://github.com/sunbus100/FreqMoE-main}{https://github.com/sunbus100/FreqMoE-main}

\end{document}